\documentclass[twoside,11pt]{article}

\usepackage{amsmath}
\usepackage{algorithm}
\usepackage{algorithmicx}
\usepackage{algpseudocode}
\usepackage{xfunctions}
\usepackage{tikz}
\usetikzlibrary{positioning}
\usepackage{pgfplots}
\usepackage{xspace}
\usepackage{url}
\usepackage{booktabs,multirow}
\usepackage{subcaption}
\usepackage[justification=centering]{caption}
\usepackage{jmlr2e}

\DeclareMathOperator*{\argmax}{argmax}
\DeclareMathOperator*{\argmin}{argmin}

\newcommand{\Fone}{F_1\xfunction}
\newcommand{\domFbeta}{{\cal D}}
\newcommand{\Fbeta}{F_\beta\xfunction}

\newcommand{\br}{\xvector{e}}

\newcommand{\brprime}{\xvector{e'}}
\newcommand{\Rnamenottobeused}{E}
\newcommand{\rpf}{\xvectorfun{\Rnamenottobeused}}
\newcommand{\hyposp}{{\cal H}}

\newcommand{\inps}{{\cal X}}
\newcommand{\Fmu}{F\xfunction}
\newcommand{\ba}{\xvectorfun{a}}
\newcommand{\bbb}{b\xfunction}

\newcommand{\bamu}{\xvectorfun{a}}

\newcommand{\calY}{{\cal Y}}
\newcommand{\prob}{\mathbb{P}}
\newcommand{\empprob}{\hat{\mathbb{P}}}
\newcommand{\paret}{\preceq}
\newcommand{\intint}[2]{\{#1,...,#2\}}
\renewcommand{\Re}{\mathbb{R}}
\newcommand{\pmu}{\mu\xfunction}

\newcommand{\domF}{{\cal D}}
\newcommand{\domFmu}{{\domF}}
\newcommand{\dotp}[2]{\left<#1,#2\right>}
\newcommand{\rpfhypo}{{\cal E}\left(\hyposp\right)}
\newcommand{\maxFmuH}{F^\star}
\newcommand{\bgg}{{\bf \hat{a}}}

\newcommand{\norm}[1]{\left\|#1\right\|_2}
\newcommand{\Lmu}{\Phi}

\newcommand{\calL}{{\cal L}}
\newcommand{\nL}{L}

\newcommand{\Pmu}{P}
\newcommand{\fpr}{\xvectorfun{{\tt FP}}}
\newcommand{\fnr}{\xvectorfun{{\tt FN}}}

\newcommand{\MFbeta}{MF_\beta\xfunction}
\newcommand{\mFbeta}{mF_\beta\xfunction}
\newcommand{\Pmuk}{P_k}
\newcommand{\tildh}{\tilde{h}}
\newcommand{\Pmupos}{1-P_1}
\newcommand{\mcFbeta}{mcF_\beta\xfunction}
\newcommand{\vecb}[1]{\boldsymbol{#1}}
\newcommand{\Jac}{Jac\xfunction}
\newcommand{\mJac}{mJac\xfunction}
\newcommand{\mcJac}{mcJac\xfunction}
\newcommand{\Prec}{Precision\xfunction}
\newcommand{\Rec}{Recall\xfunction}
\newcommand{\E}{{\mathbb E}\xfunction}
\newcommand{\tp}{\textit{tp}}
\newcommand{\tn}{\textit{tn}}
\newcommand{\fp}{\textit{fp}}
\newcommand{\fn}{\textit{fn}}

\newenvironment{rproof}[1][Proof]{\begin{trivlist} 
\item[\hskip \labelsep \textit{#1}]}{\end{trivlist}}

\jmlrheading{1}{2015}{0-0}{0/0}{0/0}{Shameem A. Puthiya Parambath, Nicolas Usunier and Yves Grandvalet}

\ShortHeadings{F-measure Optimization}{Puthiya Parambath et al.}
\firstpageno{1}

\begin{document}

\title{
Optimizing Pseudo-Linear Performance Measures: Application to F-measure
}

\author{\name{Shameem A.\ Puthiya Parambath} \email{shameem.puthiya-parambath@utc.fr} \\
       \name{Nicolas Usunier} \email{nusunier@utc.fr} \\
       \name{Yves Grandvalet} \email{yves.grandvalet@utc.fr} \\
       \addr Sorbonne universit\'es, Universit\'e de technologie de Compi\`egne -- CNRS, Heudiasyc UMR 7253\\
       Compi\`egne, France
	}
\editor{to be filled}

\maketitle

\begin{abstract}%
State of the art classification algorithms are designed to minimize the misclassification error of the system, which is a linear function of the per-class false negatives and false positives.
Nonetheless non-linear performance measures are widely used for the evaluation of learning algorithms.
For example, $F$-measure is a commonly used non-linear performance measure in classification problems.
We study the theoretical properties of a subset of non-linear performance measures called pseudo-linear performance measures which includes $F$-measure, {Jaccard index}, among many others.
We establish that many notions of $F$-measures and {Jaccard index} are pseudo-linear functions of the per-class false negatives and false positives for binary, multiclass and multilabel classification. 
Based on this observation, we present a general reduction of such performance measure optimization problem to cost-sensitive classification problem with unknown costs.
We then propose an algorithm with provable guarantees to obtain an approximately optimal classifier for the $F$-measure by solving a series of cost-sensitive classification problems.
The strength of our analysis is to be valid on any dataset and any class of classifiers, extending the existing theoretical results on binary $F$-score, which are asymptotic in nature.
Our analysis shows that thresholding cost-insensitive scores, a common technique employed to optimize $F$-measure, yields sub-optimal results. 
We also establish the multi-objective nature of the $F$-measure maximization problem by linking the algorithm with the weighted-sum approach used in multi-objective optimization.
We present numerical experiments to illustrate the relative importance of cost asymmetry and thresholding when learning linear classifiers on various $F$-measure optimization tasks.
\end{abstract}


\section{Introduction}

Different performance measures exist to assess the efficiency of learning algorithms.
Misclassification rate is the most commonly used performance measure in classification systems.
Like many other measures; which we will investigate in this paper, it is defined over the set of classification outcomes.
The four possible outcomes of a classifier are True Positive (\tp), True Negative (\tn), False Negative (\fn) and False Positive (\fp).
Misclassification rate is a linear function of these outcomes, defined as the sum of $\fp$ and $\fn$.
Conceptually, classification algorithms solve optimization problems where we optimize a loss function corresponding to the performance measure \citep[see][]{devroye1996probabilistic,anthony2009neural}.
For example, the loss function that corresponds to misclassfication rate is \emph{0-1 loss}.

As mentioned, misclassification rate is a commonly used performance measure, albeit unsuitable for specific class of problems.
For example, consider the classification (binary) of an imbalanced dataset of size 100 with 95 being samples of one specific class (let us say negative) and 5 being other class (say positive).
A trivial classifier of the form `always predict negative' results in a high accuracy albeit useless classifier.
In this specific example, $\Fbeta$ \citep{Rijsbergen79} can be considered as a more meaningful performance measure than misclassification rate.
In general, performance measures, like $\Fbeta$, are extensively used in practical problems \citep{Cheng12,bionlp13}.
One of the striking characteristics of these performance measures is the non-linearity with respect to the in-class false negatives and false positives; whereas misclassification rate is a linear function of false negatives and false positives.
Moreover, there is no convex surrogate loss function that exists for such non-linear measures; specifically, there is no surrogate loss function that exists for $F$-measure.
Another interesting property specific to $F$-measure and {Jaccard index} is: it is a sample level measure and does not decompose over individual examples.
These three aspects makes the optimization problem a difficult and interesting one.


In the current paper, we study the theoretical and algorithmic aspects pertaining to the optimization of a set of non-linear performance measures called pseudo-linear performance measures.
The commonly used performance measure $\Fone$ is an example of pseudo-linear performance measure.
Less commonly used measures like {Jaccard index} also come under this title; among many others.
Here, we focus primarily on pseudo-linear notions of $F$-measure.
We consider the setting in which a dataset, given as a set of feature vectors, is to be classified such that the $F$-measure (restricted to pseudo-linear functions) of the resulting classification is (approximately) optimal. 
In the literature, $F$-measures are also often called $F$-scores. 
Here we will stick to the first terminology, which refers to the measurement of performance, in order to avoid any confusion with classification scores, that is, the real-valued scores that may be provided by classifiers and that are thresholded to produce decisions.
Unless otherwise explicitly stated, all the discussion in this paper refers to $F$-measure optimization.
At a later point, we generalize the results to other pseudo-linear measures.

Our principle goal is to study the algorithms for empirical optimality of pseudo-linear $F$-measures.
Given a training set, our analysis proves that Optimal $F$ Classifier for pseudo-linear $F$-measures can be found by minimizing the total misclassification cost of a cost-sensitive classification for each value of cost in an inner loop and select the best among the set of costs.
Optimality in the state of the art algorithms for pseudo-linear $F$-measures are asymptotic whereas our results are valid in the non-asymptotic regime also.
Furthermore, our analysis can be linked to the weighted-sum approach used in the multi-objective optimization.
Additionally, in case of binary $\Fbeta$ and multilabel-macro-$\Fbeta$, our experimental results suggest that selecting a classifier based on minimizing the total misclassification cost is same as selecting the optimal $F$-measure \emph{a posteriori}.
Our experiments also reveals the importance of thresholding classification scores to optimize $F$-measures.

This article is an extended version of an already published conference paper \citep{Sham14}. The article is organized as follows. Section~\ref{sec:bg} introduces basic definitions and notations used throughout the paper.
It also present earlier works in $F$-measure optimization.
Section~\ref{sec:analysis} presents the theoretical analysis, where we establish the pseudo-linearity of different practical $F$-measures, and prove that Optimal $F$ Classifier can be found by minimizing the total misclassification cost of a cost-sensitive classification for a specific cost value.
We derive the values for the cost vector for many pseudo-linear $F$-measures.
We establish the multi-objective view of the $F$-measure optimization problem and link our cost-minimization approach to the popular weighted-sum approach for solving multi-objective optimization problems.
Section~\ref{sec:experiments} presents the experimental results.
We study the importance of thresholding for finding optimal solutions.
We conclude the paper in Section~\ref{sec:discus}. 
The proofs of all the propositions stated in Section~\ref{sec:analysis} are deferred to Appendix~\ref{app:thm_psd_lin}.

\section{Background and Related Work}
\label{sec:bg}
Here we give a brief review of the state-of-the-art methods for $F$-measure maximization.
We start by introducing the notations used throughout in the paper; we also give the definitions of some basic quantities like  $\Fbeta$-measure.

\subsection{Notation and Basic Definitions}

We are given \emph{(i)} a measurable space $\inps\times\calY$, where $\inps$ is the feature space and $\calY$ is the (finite) prediction set, \emph{(ii)} a probability measure $\pmu$ over $\inps\times\calY$, and \emph{(iii)} a set of (measurable) classifiers $\hyposp$ from the feature space $\inps$ to $\calY$.
We distinguish here the prediction set $\calY$ from the label space $\calL=\intint{1}{\nL}$: in binary or single-label multiclass classification, the prediction set $\calY$ is the label set $\calL$, but in multilabel classification, $\calY=2^{\calL}$ is the powerset of the set of possible labels.
In that framework, we assume that we have an i.i.d. sample drawn from an underlying data distribution $\prob$ on $\inps\times\calY$.
The empirical distribution of this finite training (or test) sample will be denoted by $\empprob$.
Then, we may take $\prob$ as measure $\pmu$ to get results at the population level (concerning expected errors), or we may take $\pmu=\empprob$ to get results on a finite sample.
Likewise, the set of classifiers $\hyposp$ can be a restricted set of functions such as linear classifiers if $\inps$ is a finite-dimensional vector space, or may be the set of all measurable classifiers from $\inps$ to $\calY$ to get results in terms of \emph{Bayes-optimal classifiers}. 
Finally, when required, we will use bold characters for vectors and normal font with subscript for indexing.

Most of the previous work on pseudo-linear metric is centered around $\Fbeta$-measure in binary settings.
$\Fbeta$-measure is defined as the weighted harmonic mean of precision and recall.
Precision is defined as the fraction of predicted positive instances that are indeed positive and recall is defined as the fraction of positive instances that are correctly predicted as positive.
Formally, we can define these metrics using classifier outcomes.
Given a binary dataset and classifier, $\tp$ corresponds to the correct prediction of a positive label, $\tn$ corresponds to the correct prediction of a negative label, $\fn$ corresponds to the incorrect prediction of a positive label as a negative label, and $\fp$ corresponds to the incorrect prediction of the negative label as positive.
In general, these outcomes are  depicted using a confusion matrix, also called contingency table (See Table~\ref{fig:conf_cost_mat}).
In terms of the classification outcomes ($\tp,\tn,\fn,\fp$), we formally define precision, recall and $\Fbeta$ associated with a binary classifier $h \in \hyposp$ for a given sample $(\vecb{x},\vecb{y}) \in (\inps\times\calY)^n$ as:

\begin{equation*}
\begin{split}
{\scriptstyle ~~~~(precision)~~~~}  \Prec{h(\vecb{x}),\vecb{y}} &= \frac{\sum_{i=1}^n\tp(h(x_i))}{\sum_{i=1}^n[\tp(h(x_i)) + \fp(h(x_i))} \\
{\scriptstyle ~~~~(recall)~~~~~~~~~~~~}  \Rec{h(\vecb{x}),\vecb{y}} &= \frac{\sum_{i=1}^n\tp(h(x_i))}{\sum_{i=1}^n[\tp(h(x_i)) + \fn(h(x_i))]} \\
{\scriptstyle ~~~~(binary-\Fbeta)~~~~~~~~~~~~} \Fbeta{h(\vecb{x}),\vecb{y}} &= \frac{(1+\beta^2)\sum_{i=1}^n\tp(h(x_i))}{\sum_{i=1}^n[(1+\beta^2) \tp(h(x_i)) + \beta^2 \fn(h(x_i)) + \fp(h(x_i))]}
\end{split}
\end{equation*}
 

In the above, dependence of label vector $\vecb{y}$ on classification outcome is omitted for convenience.
The parameter $\beta$ weights precision and recall in $\Fbeta$: $F_0$ corresponds to precision, $F_\infty$ corresponds to recall, and $\Fone$, the most widely used, corresponds to equal weights.
In case of the example mentioned in the introduction, classifying a sample of 100 instances, the trivial classifier gives precision, recall and $\Fone$ values to 0.
Precision does not consider false negatives, and recall does not consider false positives.
So in practical problems, $\Fbeta$ is preferred.
One thing to note: unlike misclassification rate, $F$-measure is not invariant under label switching i.e. if we change the positive label to negative, we get a different $F$-measure.
Hence it is used in problems where correct classification of minority label is of vital importance.
In multilabel and multiclass settings, three different definitions of $F$-measure can be found; namely instance-wise, macro and micro $F$-measures.
We will give formal definition of these in Section~\ref{sec:analysis} in connection with our theoretical framework.


\subsection{Related Work}
$F$-measure optimization had been studied on a limited basis in the past \citep{Musicant03,Jansche05,Joachims05,Jansche07,Fujino_2008}.
Last couple of years witnessed an increasing interest in this domain \citep{Dembczynski11,NanCLC12,Pillai12,Dembczynski13,Cheng12,Lipton14,NatarajanNIPS2014, HarikrishnaNIPS2014,Waegeman14a}.
Majority of the work was confined to $F$-measure maximization in binary classification settings, whereas very little work was done on multilabel and multiclass $F$-measure maximization tasks \citep{Pillai12, Dembczynski11}.
\citet{Jansche05} suggested an algorithm for finding locally maximal $\Fone$-measure for binary classification problems by approximating the classification outcomes using logistic models.
Since the objective function used is non-convex, the  algorithm does not guarantee optimality.
This issue is addressed by running the procedure multiple times and selecting the best in hand.
The orthogonal problem of infering the hypothesis with optimal $\Fone$ from a probabilistic model is discussed by (\citet{Jansche07}).
In the scientific literature, the two problem formulation has been referred to as empirical utility maximization (EUM) and decision-theoretic aproach (DTA) respectively \citep{NanCLC12}.

The two formulations differ with respect to the definition of the expected $F$-measure.
In case of the EUM based approach, population $F$-measure is defined as the $F$-measure of the expected $\tp$,$\fp$ and $\fn$.
Formally, In EUM, expected $F$-measure is defined as,
\begin{equation*}
\Fbeta^{\text{\tiny{EUM}}}(h) = \frac{(1+\beta^2)\E[\tp(h(x))]}{(1+\beta^2) \E[\tp(h(x))] + \beta^2 \E[\fn(h(x))] + \E[\fp(h(x))]}
\end{equation*}

An optimal EUM  classifier can be defined as,
\begin{equation*}
h^* = \argmax_{h \in \hyposp}\, \Fbeta^{\text{\tiny{EUM}}}(h)
\end{equation*}
In DTA, assuming a probability distribution $\vecb{p(Y)}$ on $\{0,1\}^n$, expected $F$-measure is formally defined as,
\begin{equation*}
\Fbeta^{\text{\tiny{DTA}}}(h) = \E_{\vecb{y\sim p(Y)}}[\Fbeta(h(\vecb{x}),\vecb{y})] 
\end{equation*}

An optimal DTA classifier is of the form
\begin{equation*}
h^* = \argmax_{h \in \hyposp}\, \sum_{\vecb{y} \in \{0,1\}^n}\Fbeta(h(\vecb{x}),\vecb{y})\vecb{p(y)}
\end{equation*}
From an algorithmic point of view, DTA based algorithms are computationally more expensive than EUM algorithms.
DTA based algorithms require an efficient method to estimate the joint probability and iterate over exponentially many combinations of $h$ and $y$; and the problem of estimating exact probabilities is as hard as the original problem.
But assuming i.i.d samples and considering the functional properties of $F$-measure (it is a function of integer counts ($\tp,\fp,\fn$)), the above problem can be solved more efficiently.
The algorithm given by \citet{Jansche07} runs in $O(n^4)$, where $n$ is the number of examples.
\citet{NanCLC12} improved the efficiency of this algorithm, leading to a complexity in $O(n^3)$, using dynamic programming methodology.
They also remark that the {optimal classifier} for binary $\Fone$ is of the form $sign(p(y=1|x)-\delta^*)$, where $\delta^*$ is a threshold score dependent on the underlying distribution.
\citet{Dembczynski11} extended the algorithm given by \citet{Jansche07} with dependence assumption and given a method to calculate \emph{optimal F classifier} with $O(n^3)$ complexity in time, given $n^2+1$ parameters of the joint distribution $p(\vecb{y})$.
This algorithm was used in a multilabel setting for instance-wise $F$-measure (see Remark~\ref{rem:no_psdlin}).
In addition to the high computational footprint, there is no optimality guarantee on finite samples.
In general, optimality in DTA algorithms are asymptotic in nature \citep{NanCLC12}.

On the other hand, EUM based approaches are computationally less demanding, and are based on structured risk minimization (SRM) principle. Here we minimize an approximate surrogate loss function, and select the hypothesis with minimal error on the validation set.
The most commonly employed EUM approach is to threshold the score obtained using linear classifiers like logistic regression or support vector machines (SVM) such that $\Fone$ is maximized.
An approximate surrogate function based approach named SVM\textsuperscript{perf} is given by \citet{Joachims05}, based on the observation that $\Fone$ is a sample level measure.
In the suggested method, the discriminant function is defined over the linear combination of the feature vectors, where the scalar multiplier is the label associated with each feature vector in the training sample.
Even though the reported experimental results were promising, the method does not offer any theoretical optimality guarantee.
Moreover, our experiments establish that SVM\textsuperscript{perf} is a sub-optimal method.
\citet{Musicant03} also advocated for SVMs with asymmetric costs (that is, with different costs for false negatives and false positives) for $\Fone$-measure optimization in binary classification.
However, their argument, specific to SVMs, is not methodological but technical (relaxation of the maximization problem).

In case of multilabel classification, \citet{Pillai12} argued that the multilabel-micro-$F$-measure can be optimized by thresholding the class confidence score, one label at a time.
\citet{Pillai12} used $k$-nearest neighbours and SVM to generate scores.
In general, thresholding cost-insensitive SVM scores does not guarantee empirical optimality, and the paper does not address the issue of hyperparameter selection of the backend algorithm ($k$ of $k$-nearest neighbor and regularization co-efficient of SVM).

\citet{Fujino_2008} tackle the problem by combining different classification models.
They combined two logistic models, \emph{(i)} maximum likelihood logistic regression and \emph{(ii)} approximate logistic approximation \citep[see][]{Jansche05} to maximize multilabel micro, macro and instance-wise $F$-measure.
This line of work comes under \emph{multiple classifier systems}. \emph{Multiple classifier systems} are not widely used for $F$-measure maximization, and are still in nascent stages.
In our knowledge, no proper statistical study regarding the optimality of the \emph{multiple classifier systems} for $F$-measure maximization is done so far.

Apart from $F$-measure, some of the most recent work discusses non-linear performance measures like \emph{Jaccard index} \citep{NatarajanNIPS2014, HarikrishnaNIPS2014,Waegeman14a}.
Following the footsteps of \citet{NanCLC12}, \citet{NatarajanNIPS2014, HarikrishnaNIPS2014} proposed algorithms to maximize linear-fractional performance performance measure by thresholding the class confidence score.
But as mentioned earlier, results hold only asymptotically.

In this work, we aim to perform empirical risk minimization-type learning, that is, to find a classifier with highest population level $F$-measure by maximizing its empirical counterpart. In that sense, we follow the EUM framework. Nonetheless, regardless of how we define the generalization performance, our results can be used to maximize the empirical value of the $\Fbeta$-measure.

\section{Theoretical Framework and Analysis}
\label{sec:analysis}
In this section, we present the theoretical framework which is at the heart of this work.
Our results are mainly motivated by the maximization of $F$-measures for binary, multiclass, and multilabel classification.
They rely on a general property of these performance measures, namely their pseudo-linearity with respect to the false negative and false positive probabilities.

For binary classification, we prove that, in order to optimize the $F$-measure, it is sufficient to solve a binary classification problem with different costs allocated to false positive and false negative errors (Proposition \ref{prop:weightedsum}).
However, these costs are not known \emph{a priori}, so in practice we propose to learn several classifiers with different costs, and to select the best one according to the $F$-measure in a second step.
Propositions~\ref{prop:approx} and \ref{prop:fscoresforreal} provide approximation guarantees on the $F$-measure we can obtain by following this principle depending on the granularity of the search in the cost interval.

We first establish the results for the $\Fbeta$-measures in binary classification, and then extend to other cases of $F$-measures with similar functional forms that are used in multiclass and multilabel classification.
We also briefly describe pseudo-linear notions of {Jaccard index}, which can also be solved using our framework.
For that reason, we present the results and proofs for the binary case, succeeded by multiclass and multilabel $F$-measures.

\subsection{Error Profiles and Pseudo-Linearity}

\subsubsection{Error Profiles}
The performance of a classifier $h$ on distribution $\pmu$ can be summarized by the elements of the contingency table (See Table~\ref{fig:conf_cost_mat}) which contains the summary of errors.
For all classification tasks (binary, multiclass and multilabel), the $F$-measures we consider here are functions of this non-diagonal elements of contingency table, which themselves are defined in terms of the marginal probabilities of classes and the per-class false negative/false positive probabilities. The marginal probabilities of label $k$ will be denoted by $\Pmu_k$, and the per-class false negative/false positive probabilities of a classifier $h$ are denoted by $\fnr_k{h}$ and $\fpr_k{h}$. Their definitions are given below:
\begin{equation*}
\begin{split}
{\scriptstyle (~binary/multiclass)} ~~~ & \Pmu_k = \pmu(\{(x,y)|y=k\}),~~\fnr_k{h}=\pmu(\{(x,y)|y=k\text{~and~}h(x)\neq k \})\enspace ,\\
&  \phantom{\Pmu_k = \pmu(\{(x,y)|y=k\}),~~}\fpr_k{h}=\pmu(\{(x,y)|y\neq k\text{~and~}h(x)= k \})\enspace .
\end{split} 
\end{equation*}
\begin{equation*}
\begin{split}
{\scriptstyle ~~~(multilabel)~~~~~~} ~~~ & \Pmu_k = \pmu(\{(x,y)|y\in k\}),~~\fnr_k{h}=\pmu(\{(x,y)|k\in y\text{~and~}k\not\in h(x) \})\enspace ,\\
&  \phantom{\Pmu_k = \pmu(\{(x,y)|y\in k\}),~~}\fpr_k{h}=\pmu(\{(x,y)|y\not\in k\text{~and~} k\in h(x) \})\enspace .
\end{split}
\end{equation*}

These probabilities of a classifier $h$ are then summarized by the \emph{error profile} $\rpf{h}$:
\begin{equation*}
\rpf{h}=\big(\fnr_1{h}, \fpr_1{h}, ..., \fnr_\nL{h}, \fpr_\nL{h}\big) \in \Re^{2\nL}
\enspace.
\end{equation*}




\subsubsection{Pseudo-Linear Functions}
Throughout the paper, we rely on the notion of pseudo-linearity of a function, which is itself defined from the notion of pseudo-convexity \citep[See][Definition 3.2.1]{Cambini09}: a differentiable function 
$F:\domF\subset\Re^d\rightarrow \Re$, defined on a convex open subset of $\Re^d$, is \emph{pseudo-convex} if
\begin{equation*}
\forall \br, \brprime\in \domF\enspace ,\enspace
F(\br) > F(\brprime)~~\Rightarrow ~~\dotp{\nabla F(\br)}{\brprime-\br} < 0
\enspace,
\end{equation*}
where $\dotp{.}{.}$ is the canonical dot product on $\Re^d$. 

Moreover, $F$ is \emph{pseudo-linear} if both $F$ and $-F$ are pseudo-convex.
In practice, working with gradients of non-linear functions may be cumbersome, so we will use the following characterization, which is a rephrasing of \citet[Theorem 3.3.9]{Cambini09}, basically stating that level sets of pseudo-linear functions are hyperplanes:
\begin{theorem}[\citealp{Cambini09}]
\label{thm:qualif}
A non-constant function $F\!:\domF \rightarrow \Re$, defined and differentiable on the open convex set $\domF\subseteq \Re^d$, is \emph{pseudo-linear} on $\domF$ if and only if
$\forall \br\in\domF\enspace ,\enspace \nabla F(\br) \neq \boldsymbol{0}$\enspace ,
and:
$\exists\ba\!:\Re\rightarrow\Re^d$ and $\exists\bbb\!:\Re\rightarrow \Re$ such that, for any $t$ in the image of $F$:
\begin{equation*}
F(\br) \geq t ~~\Leftrightarrow ~~\dotp{\ba(t)}{\br} +\bbb(t)\leq 0\, \text{~~~and~~~}
F(\br) \leq t~~\Leftrightarrow ~~\dotp{\ba{t}}{\br} +\bbb(t)\geq 0 \enspace .
\end{equation*}
\end{theorem}
Pseudo-linearity is the main property of linear-fractional functions (ratios of linear functions).

\begin{proposition}[Linear-fractional function]
\label{prop:fraclin}
A linear-fractional function
$F:\domF\subseteq\Re^d\rightarrow \Re$
is the ratio of linear functions, 
$F(\br) = \frac{\alpha_0 + \dotp{\boldsymbol{\gamma}}{\br}}{\alpha_1 + 
\dotp{\boldsymbol{\delta}}{\br}}$.
A non-constant linear-fractional function is pseudo-linear on the open half-space 
$\domF = \left\{\br\in\Re^d|\alpha_1+\dotp{\boldsymbol{\delta}}{\br}  > 0,~\alpha_1\neq 0\right\}$.
\end{proposition}

\subsection{Pseudo-Linearity of $F$-measures}
Several notions of $F$-measures used in practical problems are pseudo-linear.
Here, we establish that binary $\Fbeta$ and multiclass/multilabel macro/micro $F$-measures are pseudo-linear functions.

\subsubsection{Binary Classification}
\label{sec:bin_fbeta}
In binary classification, we have $\fnr_2 = \fpr_1$ and we can write $F$-measures only by reference to class $1$.
Then, for any $\beta>0$ and any binary classifier $h$, the $\Fbeta$-measure is

\begin{equation*}
\Fbeta{h}=\frac{(1+\beta^2)(\Pmu_1-\fnr_1{h})}{(1+\beta^2)\Pmu_1 + \fpr_1{h} - \fnr_1{h}}
\enspace.
\end{equation*}

\begin{figure}
\centering
    \includegraphics[width=0.9\columnwidth,trim=0mm 23cm 0mm 23cm]{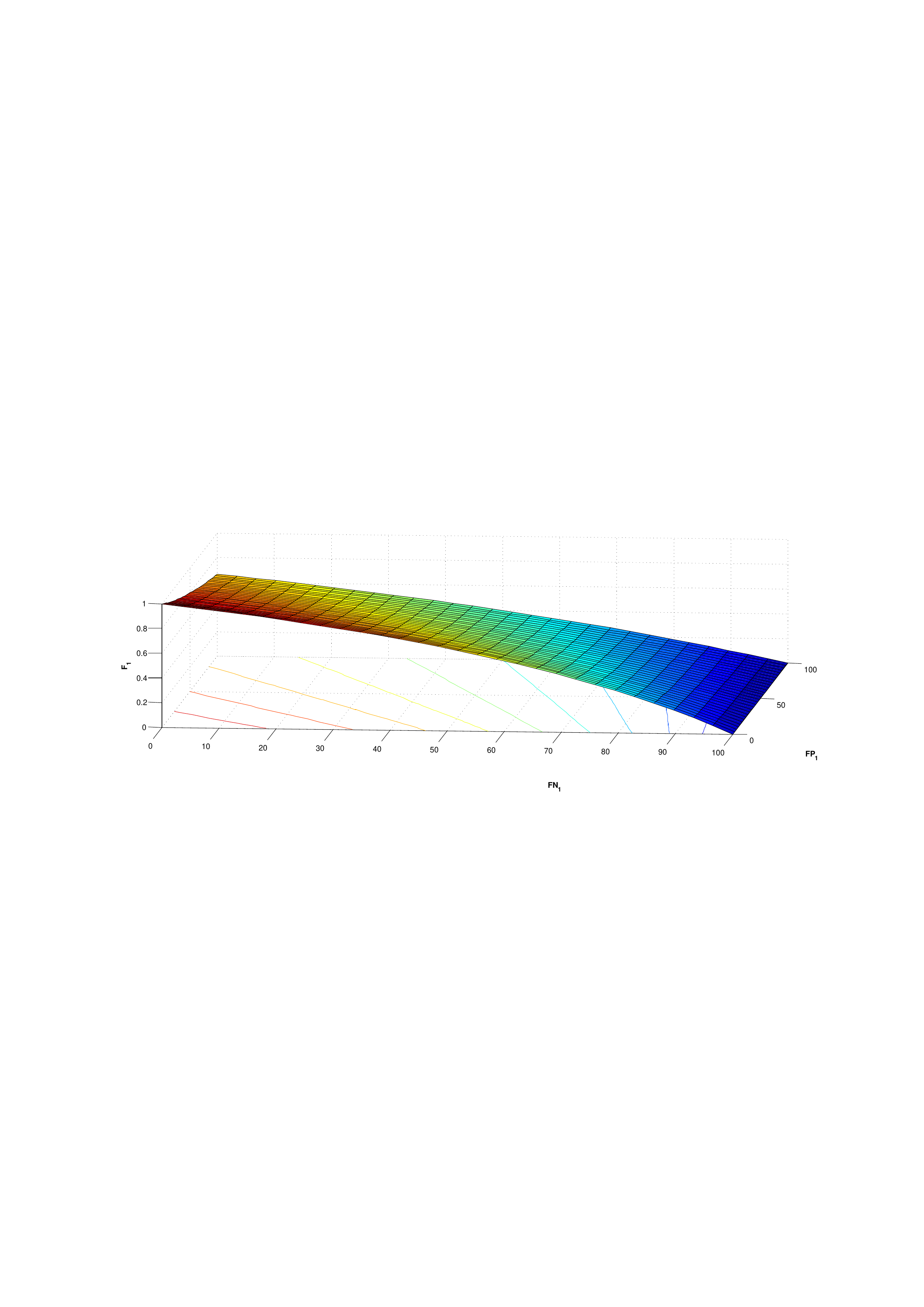}
    \caption{ Surface plot of $F_1$ as a function of $FN_1$ and $FP_1$ with level sets}
    \label{fig:bin_f1}
\end{figure}

We can immediately notice that $\Fbeta$ is linear-fractional and hence by Proposition \ref{prop:fraclin} it is pseudo-linear in $\fnr_1$ and $\fpr_1$.
Thus, with a slight (yet convenient) abuse of notation, we write the $\Fbeta$-measure for binary classification as a function of vectors in $\Re^4=\Re^{2L}$:
\begin{equation*}
{~~~~~~~~~~~\scriptstyle (binary)}\hskip \textwidth minus\textwidth \forall \br\in\Re^4, \Fbeta{\br} = \frac{(1+\beta^2)(\Pmu_1 - \br_1)}{(1+\beta^2)\Pmu_1 + \br_2 - \br_1}\, \hspace{3cm}
\end{equation*}
where $\br_i$ represents the $i^{th}$ element of the error profile $\br$. A surface plot of $\Fone$ as a function of $\fnr_1$ and $\fpr_1$ with level sets is given in Figure~\ref{fig:bin_f1}. As the Theorem~\ref{thm:qualif} states, it can be easily verified from the plot that level sets are hyperplanes.

In the above, $\br_i$ represents the $i^{th}$ element of the error profile $\br \in \rpf$.
A surface plot of $\Fone$ as a function of $\fnr_1$ and $\fpr_1$ is given in Figure~\ref{fig:bin_f1}.
It can be easily verified from the plot that level sets are hyperplanes.

\subsubsection{Multilabel Classification}
\label{sec:mullab_fbeta}
In multilabel classification, there are several definitions of $F$-measures. For those based on the error profiles, we first have the macro-$F$-measure (denoted by $\MFbeta$), which is the average over class labels of the $\Fbeta$-measure of each binary classification problem associated to the prediction of the presence/absence of a given class:
\begin{equation*}
{~~~~~~~~~~~\scriptstyle (multilabel \text{--} Macro)} \hskip \textwidth minus\textwidth \MFbeta{\br}=\frac{1}{\nL}\sum_{k=1}^\nL \frac{(1+\beta^2)(\Pmuk-\br_{2k-1})}{(1+\beta^2)\Pmuk + \br_{2k} - \br_{2k-1}}\,.\hspace{2.5cm}
\end{equation*}
$\MFbeta$ \emph{is not} a pseudo-linear function of an error profile $\br$. 
However, if the multilabel classification algorithm learns independent binary classifiers for each class \citep[a method known as one-vs-rest or binary relevance, see e.g.][]{Tsoumakas07}, then the $k$-th binary problem depends only on $\br_{2k-1}$ and $\br_{2k}$.
The maximization of the macro-$F$-measure with respect to all binary classifiers is then a separable problem which boils down to independently maximizing the $\Fbeta$-measure for $\nL$ binary classification problems.
In other words, optimizing $\MFbeta$ consists in maximizing the pseudo-linear functions in  $\br_{2k-1}$ and $\br_{2k}$ that correspond to each $\Fbeta$ optimization.
There are also micro-$F$-measures for multilabel classification.
They correspond to $\Fbeta$-measures for a new binary classification problem over $\inps\times\calL$, in which one maps a multilabel classifier $h\!:\inps\rightarrow\calY$ ($\calY$ is here the power set of $\calL$) to the following binary classifier $\tildh\!:\inps\times\calL\rightarrow\{0,1\}$: we have $\tildh(x,k)=1$ if $k\in h(x)$, and $0$ otherwise.
The micro-$\Fbeta$-measure, written as a function of an error profile $\br$ and denoted by $\mFbeta{\br}$, is the $\Fbeta$-measure of $\tildh$ and can be written as:
\begin{equation*}
{~~~~~~~~~~~\scriptstyle (multilabel \text{--} micro)} \hskip \textwidth minus\textwidth \mFbeta{\br}=\frac{(1+\beta^2)\sum_{k=1}^\nL(\Pmuk-\br_{2k-1})}{(1+\beta^2)\sum_{k=1}^\nL\Pmuk +\sum_{k=1}^\nL(\br_{2k}- \br_{2k-1})}\enspace .\hspace{1.75cm}
\end{equation*}
This function is also linear-fractional, and thus pseudo-linear in $\br$.

\subsubsection{Multiclass Classification}
\label{sec:mulclass_fbeta}
The last example we take is from multiclass classification. It differs from multilabel classification in that a single class must be predicted for each example.
This restriction imposes strong global constraints that make the multiclass classification significantly harder.
As for the multilabel case, there are many definitions of $F$-measures for multiclass classification, and in fact several definitions for the micro-$F$-measure itself.
We will focus on the following one, which is used in information extraction \citep[e.g in the BioNLP Challenge][]{bionlp13}.
Given $\nL$ class labels, we will assume that label $1$ corresponds to a ``default'' class, the prediction of which is considered as not important.
In information extraction, the default class corresponds to the (majority) case where no information should be extracted.
Then, a false negative is an example $(x,y)$ such that $y\neq 1$ and $h(x)\neq y$, while a false positive is an example $(x,y)$ such that $y = 1$ and $h(x)\neq y$. 
This micro-$F$-measure, denoted $\mcFbeta$ can be written as:
\begin{equation*}
{~~~~~~~~~~~\scriptstyle (multiclass \text{--} micro)} \hskip \textwidth minus\textwidth \mcFbeta{\br} = \frac{(1+\beta^2)(\Pmupos - \sum_{k=2}^\nL \br_{2k-1})}{(1+\beta^2)(\Pmupos) - \sum_{k=2}^\nL\br_{2k-1}+\br_1}\enspace . \hspace{2cm}
\end{equation*}
Once again, this kind of micro-$\Fbeta$-measure is linear-fractional and hence pseudo-linear in $\br$.

\begin{remark}[Non-pseudo-linear F-measures]
\label{rem:no_psdlin}
In multilabel settings, notion of instance-wise $\Fbeta$ has been used in the past \citep{Fujino_2008,Dembczynski11,petterson2010,petterson2011,Cheng12,Dembczynski13}.
It is similar to the micro-$F$-measure ($\mFbeta$) for multilabel case defined above, but defined over samples (instances) instead of labels.
It is defined as the average of the per-instance $F$-measure.
Hence, we calculate the $F$-measures for each instance independently (i.e. estimate $\mFbeta$ for each individual example by calculating $\tp,\fp,\fn$ for each example in the sample) and take the average (arithmetic mean) over the number of samples.
This measure can not be written as a linear-fractional function of ``error profile'' terms, hence it can not be solved using our framework.
\end{remark}

\subsection{Optimizing $F$-Measure by Reduction to Cost-Sensitive Classification}
The $\Fbeta$-measures presented above are non-linear aggregations of false negative/positive propotions that can not be written in the usual expected loss minimization framework; usual learning algorithms are thus, intrinsically, not designed to optimize this kind of performance measures.
We show in Proposition \ref{prop:weightedsum} that the optimal classifier for a cost-sensitive classification problem with label dependent costs \citep{Elkan2001,Zhou2010} is also an optimal classifier for the pseudo-linear $F$-measures (within a specific, yet arbitrary classifier set $\hyposp$). 
In cost-sensitive classification,  each entry of the error profile is weighted asymmetrically by a non-negative cost, and the goal is to minimize the weighted average error.
Efficient, consistent algorithms exist for such cost-sensitive problems \citep{Abe04,Steinwart07,Scott12}.
Even though the costs corresponding to the optimal $F$-measure are not known \emph{a priori}, we show in Proposition \ref{prop:approx} that we can approximate the optimal classifier with approximate costs.
These costs, explicitly expressed in terms of the optimal $F$-measure, motivate a practical algorithm.
Even though the discussion in this section is more general and applies to any pseudo-linear functions, we start with the discussion in binary settings.
We give the proofs and results for binary $\Fbeta$ and extend the results to multilabel and multiclass $F$-measures in Section~\ref{sec:beyond_fbeta}.

\subsubsection{Reduction to Cost-Sensitive Classification}
\label{sec:red_cs}
Let $\Fmu:\domFmu\subset\Re^d\rightarrow \Re$ be a fixed pseudo-linear function.
We denote by $\bamu:\Re\rightarrow \Re^d$ the function mapping values of $\Fmu$ to the corresponding level set of Theorem~\ref{thm:qualif}.
We assume that the distribution $\pmu$ is fixed, as well as the (arbitrary) set of classifier $\hyposp$.
We denote by $\rpfhypo$ the closure of the image of $\hyposp$ under $\rpf$, i.e. $\rpfhypo = cl(\{\rpf{h}, h\in\hyposp\})$ (the closure ensures that $\rpfhypo$ is compact and that minima/maxima are well-defined), and we assume $\rpfhypo\subseteq \domFmu$. Finally, for the sake of discussion with cost-sensitive classification, we assume that $\bamu{t} \in \Re_+^d$ for any $\br \in\rpfhypo$, that is, lower values of errors entail higher values of $\Fmu$.

\begin{proposition}
\label{prop:weightedsum}
Let ${\displaystyle \maxFmuH = \max_{\br\in\rpfhypo} \Fmu{\br}}$. We have:~
$\displaystyle \br^\star \in \argmin_{\br\in\rpfhypo} \dotp{\bamu\big(\maxFmuH\big)}{\br}~\Leftrightarrow~\Fmu{\br^\star}=\maxFmuH$.
\end{proposition}

This proposition shows that $\bamu\big(\maxFmuH\big)$ are the cost vectors, which are orthogonal to the level set of $\Fmu$ at $\maxFmuH$ and may not need to be unique, that should be assigned to the error profile in order to find the optimal classifier in $\hyposp$ with respect to the measure $\Fmu$.
Hence maximizing $\Fmu$ amounts to minimizing $\dotp{\bamu\big(\maxFmuH\big)}{\rpf{h}}$ with respect to $h$, that is, amounts to solving a cost-sensitive classification problem.
This observation suggests that the optimization of pseudo-linear measures could be a wrapper of cost-sensitive classification algorithms. 
The costs $\bamu\big(\maxFmuH\big)$ are, however, not known \emph{a priori}. 
The following result shows that having only approximate costs is sufficient to have an approximately optimal solution, which gives us the main step towards a practical solution.

\begin{proposition}
\label{prop:approx}
Let $\varepsilon_0\geq 0$ and $\varepsilon_1\geq 0$, and assume that there exists $\Lmu>0$ such that for all $\br, \brprime\in\rpfhypo$ satisfying $\Fmu{\brprime}>\Fmu{\br}$, we have:
\begin{equation*}  
\Fmu{\brprime}-\Fmu{\br}\leq \Lmu\dotp{\bamu{\Fmu(\brprime)}}{\br-\brprime}  \enspace.  
\end{equation*}
Then, let us take $\br^\star\in\argmax_{\brprime\in\rpfhypo} \Fmu{\brprime}$, and denote $\ba^\star=\bamu{\Fmu(\br^\star)}$. Let furthermore $\bgg\in\Re_+^d$ and $h\in\hyposp$ satisfying the following conditions:
\begin{center}
{(i)} $\norm{\bgg-\ba^\star}\leq \varepsilon_0$
\enspace,
\hspace{2.5cm} {(ii)} $\displaystyle \dotp{\bgg}{\br} \leq \min_{\brprime\in\rpfhypo} \dotp{\bgg}{\brprime}+\varepsilon_1$
\enspace.
\end{center}
We have: $\forall \br \in \rpfhypo, ~ \Fmu{\br}\geq \Fmu{\br^\star} - \Lmu \cdot (2\varepsilon_0M+\varepsilon_1)$
\enspace,
\enspace where $\displaystyle M=\max_{\brprime\in\rpfhypo}{\norm{\brprime}}$
\enspace. 
\end{proposition}

The above proposition suggests that pseudo-linear measures could be optimized by wrapping cost-sensitive classification in an inner loop with an outer loop setting the appropriate costs.
This proposition also gives an upper bound on the achievable optimal $F$-score.
This value depends on the size of the maximum error associated with the given hypothesis space,$M$, measured in $\ell_2$ sense and the constant $\Lmu$.
The value of $M$ depends on the selected hypothesis class ($\rpfhypo$). 
We call $\Lmu$ as discretization factor as it defines the granularity of the approximation.
It depends on the specific form of $F$-measure and training sample.
We can find an approximately optimal classifier using a procedure, where we search for an approximately optimal cost and associated error profile by iterating through the preselected cost interval in small steps.
Thus searching for a cost such that $\varepsilon_0$ is close to zero, we can find an approximately \emph{optimal F classifier}.
$\varepsilon_1$ can be regarded as the approximation guarantee provided by the underlying cost-sensitive classification algorithm.
Practical implementations use convex surrogate loss instead of the non-convex \emph{0-1} loss.
A discussion on convex approxmiation of \emph{0-1 loss} can be found in \citep{Rosasco_areloss}.
$\Lmu$, the discretization factor gives the magnitude of the step size.
A larger value of $\Lmu$ indicates more fine-grained discretization (very small step size), and a smaller value of $\Lmu$ indicates coarse- grained discretization.
 Later, we will derive the exact values of $\Lmu$ and the cost interval for specific $F$-measures.

\subsubsection{Discretization Factor  and Cost Interval for $F_{\beta}$}
Here, we derive the values of the discretization factor ($\Lmu$) and the range of the cost interval ($\ba$) for binary $\Fbeta$-measure.

\begin{proposition}
\label{prop:fscoresforreal}
$\Fbeta$ defined in Section \ref{sec:bin_fbeta} satisfy the conditions of Proposition \ref{prop:approx} with:
\begin{equation*}
{\scriptstyle ~~~~~~~~~~(binary)~~\Fbeta:} \hspace{1.6cm} \!\Lmu=\frac{1}{\beta^2\Pmu_1} \hspace{0.6cm} \text{~~~and~~~~~~~~} \bamu:t\in[0,1]\mapsto(1+\beta^2-t,t,0,0)\enspace .\hspace{1.6cm}
\end{equation*}
\end{proposition}
This proposition gives the exact values of $\Lmu$ and the range for $\ba$ in binary settings.
Here the discretization factor depends on the marginal probability of the positive class (assume label 1 represents positive class).
A larger value of the discretization factor demands smaller step size in the cost interval.
Looking at the approximation guarantee in proposition~\ref{prop:approx}, with a larger value of $\Lmu$, reasonable approximation can be obtained by taking $\varepsilon_0$ close to zero.
Intuitively, we can think of this as follows,
higher values of $\Lmu$ indicates a highly imbalanced data with very few positive examples, hence to eliminate the influence of class-imbalance, we need to discretize in smaller step through cost interval.
Given the error profile (in the form of contingency table) and associated costs as a matrix, as shown in in Figure~\ref{fig:conf_cost_mat}, corresponding $\Fbeta$-measure is the sum of the elements of the Hadamard product of the two matrices.

\newcommand{\confmatrix}{
\matrix (conmat) [ampersand replacement = \&] {
\node (tpos) [box, align = center, 
    label=left:\( \mathbf{P} \),
    label=above:\( \mathbf{P} \)
    ] {True  Positive \\ (\tp)};
\&
\node (fneg) [box, align = center,
    label=above:\textbf{N},
	] {False  Negative \\ (\fn)};
\\
\node (fpos) [box, align = center,
    label=left:\( \mathbf{N} \),
	] {False  Positive \\ (\fp)};
\&
\node (tneg) [box, align = center,
	] {True  Negative \\ (\tn)};
\\
};
\node [rotate=90,left=.01cm of conmat, text width=3.2cm,align=left,anchor=center] {\textbf{Actual Label}};
\node [above=.01cm of conmat, align=center, anchor=center] {\textbf{Predicted Label}};
}

\newcommand{\costmatrix}{
\matrix (costmat) [row sep=.01cm,column sep=.01cm, ampersand replacement = \&] {
\node (tpcos) [box, align = center] {$0$};
\&
\node (fncos) [box, align = center] {$1+\beta^2-t$};
\\
\node (fpcos) [box, align = center] {$t$};
\&
\node (tncos) [box, align = center] {$0$};
\\
};
}

\begin{figure}
\captionsetup[subfigure]{font=footnotesize}
\centering
\subcaptionbox{Contingency Table}[.4\textwidth]{
\begin{tikzpicture}[box/.style={draw,rectangle,minimum width=2.8cm,minimum height=1.4cm,align=left}]
\confmatrix
\end{tikzpicture} }
\subcaptionbox{Cost Matrix}[.4\textwidth] {
\begin{tikzpicture}[box/.style={draw,rectangle,minimum width=2cm,minimum height=1.4cm,align=left}]
\costmatrix
\end{tikzpicture} }
\caption{Binary Classification}
\label{fig:conf_cost_mat}
\end{figure}

\begin{corollary}
\label{coro:foptcost}
{
For the $\Fone$-measure, the optimal classifier is the solution to the cost-sensitive binary classifier with costs $\big(1-\frac{\maxFmuH}{2},\frac{\maxFmuH}{2}\big)$
} 
\end{corollary}
This proposition extends the result obtained by \cite{Lipton14} to the non-asymptotic regime.
If we take $\hyposp$ as the set of all measurable functions, the Bayes-optimal classifier for this cost is to predict class 1 when $\pmu(y=1|x) \ge \frac{\maxFmuH}{2}$ \citep[see][]{Lipton14,Steinwart07}.
\subsubsection{Algorithm for $F_\beta$ Maximization}
Based on the above results, we give a practical algorithm to find optimal $\Fbeta$.
In case of $\Fbeta$, the cost function $\bamu:[0,1]\rightarrow \Re^d$, which assigns costs to probabilities of error, is Lipschitz-continuous with Lipschitz constant ($\phi) = \max(1, \beta^2)$.
Hence it is sufficient to discretize the interval $[0,1]$ to have a set of evenly spaced values $\{t_1, ..., t_C\}$ (say, $t_{j+1}-t_j=\varepsilon_0/\phi$) to obtain an $\varepsilon_0$-cover $\{\bamu(t_1), ..., \bamu(t_C)\}$ of the possible costs.
Using the approximate guarantee of Proposition \ref{prop:approx}, learning a  cost-sensitive classifier ($h_i$) for each $\bamu(t_i)$ and selecting the one with minimum total misclassification cost($\dotp{\bamu(t_i)}{h_i(\br)}$) on a validation set is sufficient to obtain a $\Lmu(2\varepsilon_0M+\varepsilon_1)$-optimal solution. 
Our experimental results suggest that, in binary classification choosing a classifier by our proposed method is same as selecting a classifier with optimal $F$-measure \emph{a posteriori}. Hence our final algorithm consists of selecting a cost-sensitive classifier with optimal $F$-score.
Our suggested algorithm is presented in Algorithm~\ref{alg:optfb}.

\begin{algorithm}
\caption{Optimization of the $F_\beta$-measure}
\label{alg:optfb}
\begin{algorithmic}[1]
\Procedure{Optimize\_$F_\beta$}{D,$\beta$} \Comment {D = Data, $\beta = \beta$ in $F_{\beta}$}
\State{$bF$ = 0}
\State{Split Training Data into two $D_{tra},D_{val}$ }
\For{$t = (0\dotsc1+\beta^2)$} \Comment{approximate cost}
\State{$\phi,\theta,F$ = F\_cs\_learner($D_{tra},D_{val},t)$}; \Comment{learn cost-sensitive model}
\If{$F > bF$}
\State{$\Phi = \phi, ~\Theta = \theta, ~bF = F$; }
\EndIf
\EndFor
\State \Return ($\Phi,~\Theta$)
\EndProcedure
\end{algorithmic}
\end{algorithm}

\begin{algorithm}
\caption{Cost-Sensitive Learner for $F_\beta$}
\label{alg:cs_fb}
\begin{algorithmic}[1]
\Procedure{F\_cs\_learner}{$D_{tra},D_{val},t$} \Comment {$D_{tra}$ = Training Data, $D_{val}$ = Validation Data, t=cost}
\State{$bF$ = 0}
\For{$\psi \in \Psi$} \Comment {$\Psi$ = set of tunable cost-sensitive algorithm hyper-parameter}
\State{$\phi$ = cost\_sensitive\_learner($D_{tra},t,\psi$)}; \Comment{generic cost-sensitive learner}
\State{$\theta,F$= compute$\Fbeta$($\phi,D_{val},\beta$)} \Comment{get optimal threshold and $\Fbeta$}
\If{$F > bF$}
\State{$\Phi = \phi, ~\Theta = \theta, ~bF = F$; }
\EndIf
\EndFor
\State \Return ($\Phi,~\Theta,~F$)
\EndProcedure
\end{algorithmic}
\end{algorithm}

The cost-sensitive classification algorithms that are used in the inner loop (step 5) returns the trained model.
The $predict\_score$ method in the meta-algorithm simply returns the scores (score can be posterior probability, or geometric margin etc) on the validation set and $computeF_\beta$ returns the optimal $F$-measure and a score threshold (if any) on the validation data.
Even though our theoretical results do not suggest thresholding the scores \emph{a posteriori}, experimental results indicate the need for a posterior thresholding of the scores. We will elaborate on this point in Section~\ref{sec:experiments}.
This meta-algorithm can be instantiated with any cost-sensitive learning algorithm.
The actual algorithm may simply consist of adjusting the hyper-parameters of a cost-insensitive classifier so as to optimize cost-sensitive classification, as in many practical implementation of cost-sensitive algorithm.
This rudimentary approach results in considerable savings in computation time.

\subsection{Beyond Binary $F$-measure}
\label{sec:beyond_fbeta}
As mentioned earlier, many notions of $F$-measures in multiclass and multilabel problems are pseudo-linear and can be solved using our framework.
Here, we derive the values of cost vector ($\bamu$) and discretization factor ($\Lmu$), and give optimal $F$-measure algorithm for pseudo-linear $F$-measures described in Sections~\ref{sec:mullab_fbeta} and \ref{sec:mulclass_fbeta}.

\subsubsection{Multilabel micro-$F$-measure}

\begin{proposition}
\label{prop:mfscoresforreal}
{
multilabel micro-$F$($\mFbeta$) defined in Section~\ref{sec:mullab_fbeta} satisfies the conditions of Proposition \ref{prop:approx} with:
\begin{equation*}
{\scriptstyle (multilabel \text{--} micro)~~\mFbeta:} \hspace{0.41cm} \,\Lmu=\frac{1}{\beta^2\sum_{k=1}^\nL \Pmuk} \hspace{0.15cm} \text{~~and~~} \bamu_i(t)=\begin{cases}1+\beta^2-t& \text{~if $i$ is odd}\\ t &\text{~if $i$ is even}\end{cases}\enspace .\hspace{1.75cm}
\end{equation*}
}
\end{proposition}

Here the discretization factor depends on the sum of marginal probabilities of each label.
A large value of $\Lmu$ indicates that majority of the labels are rare, and smaller value of $\Lmu$ indicates that few labels are rare.
Since the impact of misclassification of rare labels does not influence the micro-$F$-measure to a greater extend ($F$-score is independent of true negatives), we have to discretize in a smaller step only if the majority of the classes are rare.
Given the above result on cost vector $\bamu$ and discretization factor $\Lmu$, and following the arguments given for $\Fbeta$ (here also the cost function $\bamu$ is Lipschitz-continuous with Lipschitz constant taking value $max(1,\beta^2)$), we can develop an algorithm for finding optimal classifier for $\mFbeta$.
Unlike in binary case, here we run cost-sensitive learner with discretized cost values to find the classifier with lowest total misclassification cost($\dotp{\bamu(t_i)}{h_i(\br)}$).
Our proposed algorithm is given in Algorithm~\ref{alg:optmf}.
The algorithm is similar to the $\Fbeta$ algorithm given in Algorithm~\ref{alg:optfb}, except for the fact that here we minimize the total misclassification cost instead of maximixing empirical $\Fbeta$ in the inner loop.
Also, here we need the cardinality of the label space as an additional input parameter.
Here the outer loop calculates the cost ($\bamu(t)$) for each value of $t$ as given in proposition~\ref{prop:mfscoresforreal}.
The selected threshold is the one which minimizes the total misclassification cost ($\dotp{\bamu(t)}{\br}$) over all possible values of $\bamu(t)$ and $\br$.

\begin{algorithm}
\caption{Optimization of the $mF_\beta$-measure}
\label{alg:optmf}
\begin{algorithmic}[1]
\Procedure{Optimize\_$mF_\beta$}{D,L,$\beta$} \Comment {D = Data, L = $\lvert \calL \lvert$, $\beta = \beta$ in $F_{\beta}$}
\State{$bC = +\infty$}
\State{$bmF = 0$}
\State{Split Training Data into two $D_{tra},D_{val}$ }
\For{$t = (0\dotsc1+\beta^2)$} \Comment{Approximate Cost}
\State{$\Pi$ = gen\_$mF_\beta$\_cost\_vector($L,t,\beta$)} \Comment{Cost Vector}
\State{$\phi,\theta$ = mF\_cs\_learner($D_{tra},D_{val},\Pi$)} \Comment{learn cost-sensitive model}
\State{$\theta,mF$ = compute$\mFbeta$($\phi,D_{val},\theta,\beta)$} \Comment{get the optimal threshold and $\mFbeta$}
\If{($mF > bmF$)}
\State{$bmF = mF,~ \Phi = \phi,~ \Theta = \theta$; }
\EndIf
\EndFor
\State \Return ($\Phi,\Theta$)
\EndProcedure
\end{algorithmic}
\end{algorithm}

\begin{algorithm}
\caption{Cost-Sensitive Learner for $mF_\beta$}
\label{alg:cs_mfb}
\begin{algorithmic}[1]
\Procedure{mF\_cs\_learner}{$D_{tra},D_{val},\Pi$} \Comment {$D_{tra}$ = Training Data, $D_{val}$ = Validation Data, $\Pi$=cost}
\State{$bC = +\infty$}
\For{$\psi \in \Psi$} \Comment {$\Psi$ = set of tunable cost-sensitive algorithm hyper-parameter}
\State{$\phi$ = cost\_sensitive\_learner($D_{tra},\Pi,\psi$)}; \Comment{generic cost-sensitive learner}
\State{$\theta,~C$= compute\_cost($\phi,D_{val},\Pi$)} \Comment{get optimal threshold and total misclassification cost}
\If{($C < bC$)}
\State{$\Phi = \phi, ~\Theta = \theta$; }
\EndIf
\EndFor
\State \Return ($\Phi,~\Theta$)
\EndProcedure
\end{algorithmic}
\end{algorithm}


\subsubsection{Multiclass micro-$F$-measure}

\begin{proposition}
\label{prop:mcfscoresforreal}
multiclass micro-$F$($\mcFbeta$) defined in Section~\ref{sec:mulclass_fbeta} satisfies the conditions of Proposition \ref{prop:approx} with:
\begin{equation*}
{\scriptstyle (multiclass \text{--} micro)~~\mcFbeta:} \hspace{0.37cm} \!\Lmu=\frac{1}{\beta^2(\Pmupos)} \hspace{0.31cm} \text{~~and~~} \bamu_i (t)=\begin{cases}1+\beta^2-t& \text{~if $i$ is odd and $i\neq 1$}\\ t &\text{~if $i=1$}\\0&\text{~otherwise}\end{cases}\enspace .\hspace{0.2cm}
\end{equation*}
\end{proposition}

Following the arguments given for multilabel micro-$F$-measure, we can use the Algorithm~\ref{alg:optmf} for finding optimal $\mcFbeta$ with a small modification to the $gen\_mF_\beta\_cost\_vector$ method. The new cost generation method for multiclass micro-$F$-measure follows result of proposition~\ref{prop:mcfscoresforreal}.


\begin{remark}[Beyond $F$-Measures]
Jaccard index is a set-based similarity measure.
Given two sets, Jaccard index is defined as the ratio of intersection to union.
Like $\Fone$-measure, it ranges from $0$ to $1$, where $0$ indicates distinct sets and $1$ indicates identical sets (\cite{kaufman2009finding}).
It is used in cluster analysis and co-citation analysis to name a few.
Some recent work (\citep{Waegeman14a,NatarajanNIPS2014}) examined the use of Jaccard index as a performance measure in classification problems.
The Jaccard index is a pseudo-linear performance function of per-class false negatives and false positives.
We can define Jaccard indexes for binary, multiclass and multilabel problems in terms of
the \emph{error profile} entries,
\begin{equation*}
\begin{split}
\hspace{2cm}{\scriptstyle (binary)}\hspace{2.2cm} & \forall \br\in\Re^4, \quad \Jac{\br}=\frac{\Pmu_1-\br_1}{\Pmu_1+\br_2}\,\hspace{3cm} \\
\hspace{2cm}{\scriptstyle (multilabel \text{--} micro)} \hspace{2.2cm} & \forall \br\in\Re^{2L}, \quad \mJac{\br}=\frac{\sum_{k=1}^\nL(\Pmuk-\br_{2k-1})}{\sum_{k=1}^\nL\Pmuk +\sum_{k=1}^\nL \br_{2k}}\, \hspace{1.75cm} \\
\hspace{2cm}{\scriptstyle (multiclass \text{--} micro)} \hspace{2.2cm} & \forall \br\in\Re^{2L}, \quad \mcJac{\br}=\frac{\Pmupos - \sum_{k=2}^\nL \br_{2k-1}}{(\Pmupos) +\br_1}\, \hspace{1.75cm}\\
\end{split}
\end{equation*}
As we can infer from the above equations, these quantities are pseudo-linear and hence, we can use the methodology developed in Section~\ref{sec:red_cs},thresholding cost-sensitive scores, to find optimal Jaccard index classifier. Our analysis proves the remark of \cite{Waegeman14a} ``We also see that algorithms maximizing the F-measure perform the best for Jaccard index". 
\end{remark}

\section{Relationship to Multi-Objective Optimization}
\label{sec:mult_obj}

Finding ``good'' classifiers amounts to find good trade-offs between the different types of errors. In any case, it is a natural requirement that the chosen classifier has an error profile that is a minimal element of $\rpfhypo$ according to the partial order of Pareto dominance, which is denoted by $\paret$ and is defined as:

\begin{equation*}
  \forall \br, \brprime\in\Re^d\enspace ,\enspace 
  \br \paret \brprime ~~\Leftrightarrow~~ \forall k\in\intint{1}{d}\enspace , \enspace
  \br_k\leq\brprime_k
  \enspace.
\end{equation*}
The set of optimal solutions defines the Pareto front.

error profile that is a minimal element of $\rpfhypo$ according to Pareto-dominance
 (where $\br \succeq \brprime$ iff $\br_k\geq\brprime_k$ for all $k$). This set of optimal solutions defines the Pareto front.

Multi-objective optimization defines methods for finding the Pareto front, or approximations of it (\cite{EhrgottGandibleuxbook02}), and one of the motivations is to find (approximately) optimal solutions of a vector function that is hard to optimize.
The process is to generate candidate points in the Pareto front, and take the candidate with optimal value of the vector function.
The advantage is generating candidate points is faster than the direct optimization of the vector function. In our case, goal is to find $h \in \rpfhypo$ that achieves small values of $\dotp{\ba}{\br(h)}$ for a predefined cost vector $\ba$.

The reduction from pseudo-linear functions to solving a series of cost-sensitive classification problems exactly corresponds to this Pareto front method.
In fact, a general way of finding Pareto-optimal solutions of a multi-objective problems is called the weighted-sum method (see e.g. \cite{EhrgottGandibleuxbook02,Boyd:2004:CO:993483}).
Applied to error profiles, the weighted-sum method would minimize positive weighted combinations of the elements of the error profiles, which corresponds to solving a cost-sensitive classification problem.
In usual multi-objective optimization settings, such a Pareto set method is not useful for pseudo-linear aggregation functions, because most such functions are linear-fractional, and single-objective problems with a linear-fractional objective function can be rewritten in terms of a linear objective with linear constraints (see e.g. \cite{Boyd:2004:CO:993483}).
In our context however, the linearization would not help because it would introduce constraints involving values of the error profiles, which are not linear in general.
What we gain with the reduction to cost-sensitive classification (or, equivalently, with the weighted-sum method), is that efficient algorithms for cost-sensitive classification, which are known to work in practice and are asymptotically optimal, are already known.
In addition, weighted-sum method require the users to know the relative preferences of the objectives in advance, which is not known in general. Hence the weight components are unbounded. Our reduction clearly defines a bound on the possible weights ($\ba(t)$).

\newcommand{\paretofrontfig}{%
\addplot[red,mark=none] coordinates {
(0.0,0.5825)
(0.195,0.1275)
(0.375,0.0075)
(0.4175,0.0)
};
\addlegendentry{Convex hull}
\addplot[dashed,blue,mark=+,opacity=0.5] coordinates {
(0.0,0.5825)
(0.0425,0.575)
(0.18,0.4625)
(0.195,0.1275)
(0.2375,0.12)
(0.375,0.0075)
(0.4175,0.0)
};
\addlegendentry{Pareto front}
\node[above right,blue] at (axis cs:0.0,0.5825) {$c_A$};
\node[above right,blue] at (axis cs:0.0425,0.575) {$c_B$};
\node[above ,blue] at (axis cs:0.18,0.4625) {$c_C$};
\node[above ,blue] at (axis cs:0.195,0.1275) {$c_D$};
\node[above ,blue] at (axis cs:0.2375,0.12) {$c_E$};
\node[above ,blue] at (axis cs:0.375,0.0075) {$c_F$};
\node[above ,blue] at (axis cs:0.4175,0.0) {$c_G$};
}
\newcommand{\paretofronttable}{%
$h_A(x)$ & $2$ & $2$ & $2$ & $2.22$ \\
$h_B(x)$ & $2$ & $2$ & $1$ & $2.37$ \\
$h_C(x)$ & $2$ & $1$ & $2$ & $27.22$ \\
$h_D(x)$ & $1$ & $2$ & $2$ & $73.83$ \\
$h_E(x)$ & $1$ & $2$ & $1$ & $72.12$ \\
$h_F(x)$ & $1$ & $1$ & $2$ & $75.24$ \\
$h_G(x)$ & $1$ & $1$ & $1$ & $73.62$ \\
}
\newcommand{\paretofronttablelegend}{%
classifier & $x_0$ & $x_1$ & $x_2$ & $F_1^\mu$ (\%) \\
}
\newcommand{\paretofrontdata}{%
$\mu(x)$ & 0.65 & 0.30 & 0.05 \\
$\mu(y=1|x)$ & 0.70 & 0.40 & 0.15 \\
}
\newcommand{\paretofrontdatalegend}{%
  & $x_0$ & $x_1$ & $x_2$ &   \\
}

\begin{figure}[t]
\begin{tikzpicture}[scale=0.8]
\begin{axis}[xmin=0,xmax=0.5,ymin=0,ymax=0.65,xlabel={proportion of false positives}, ylabel={proportion false negatives},legend style={at={(0.98,0.7)},anchor=east}]
\paretofrontfig
\end{axis}
\end{tikzpicture}
\hfill
\begin{tabular}[b]{ccccc}
\hline
\paretofrontdatalegend
\hline
\paretofrontdata
\hline
\paretofronttablelegend
\hline
\paretofronttable[5ex]
& & & & \\
\end{tabular}
\caption{Pareto front for a binary classification problem ($\calY=\{1, 2\}$, the positive class is $1$), where the input space contains three points $x_1$, $x_2$, $x_3$. The table on the left describes the data distribution, and defines the 8 possible classifiers and gives their $F_1^\mu$-measure.}
\label{fig:paret}
\end{figure}

The relationship between the reduction to cost-sensitive classification and the weighted-sum method allows us to discuss pseudo-linear F-measures in terms of Pareto-optimal solutions.
It is well-known that in general, not all Pareto-optimal solutions can be found by the weighted-sum method; in fact, only those that are on the boundary of the \emph{convex hull} of the feasible set can be reached.
In general however, many classification problems have Pareto-optimal solutions that do not lie on this boundary, especially if the input space is finite (as is the case on any finite dataset).
Figure \ref{fig:paret} gives the example of the Pareto front of a binary classification problem with 3 examples.
The pareto front can be depicted on a 2D plane where the axis are false positives and false negatives; up to a change of basis, this Pareto front is the ROC curve \citep{Bach06,clemenccon09} for the problem.
In the figure, the blue points on the left plot correspond to Pareto-optimal classifiers (none of them can be improved both in terms of proportion of false positives and false negatives), while the red curve is the Pareto set of the convex hull of the error profiles of the $8$ classifiers. Our result of reduction to cost-sensitive classification proves that only the classifiers whose error profile is both Pareto-optimal and on the boundary of the convex hull are candidates as optimal classifiers for any pseudo-linear aggregation function (here, the candidates are $c_A, c_D, c_F$), even though all classifiers are optimal for some trade-off rule. For instance, $c_B$ is the optimal classifier for the rule "`minimize the proportion of false negatives under the constraint that the proportion of false positives is smaller than 0.1"'.

\section{Experiments}
\label{sec:experiments}

This section illustrates of the accuracy of the algorithms suggested by our theoretical framework, using the $\Fone$-measure, in binary and multilabel classification.
Our experimental results for binary and multilabel-macro $F$-measure (using binary relevance) shows that \emph{(i)} choosing Optimal F Classifier by minimizing $\dotp{\ba}{\br}$ is same as choosing classifier with optimal $F$-measure \emph{a posteriori}
(\emph{ii}) selecting a classifier by thresholding cost-sensitive scores is preferable to algorithms based on thresholding cost-insensitive classification scores: to maximize $F$-measure
(\emph{iii}) In case of multilabel-micro $F$-measure, Optimal F Classifier is the one with lowest $\dotp{\ba}{\br}$ value.

We compare thresholded cost-sensitive classification, as implemented by SVMs and logistic regression (LR), with asymmetric costs, to thresholded linear classifiers (SVMs and logistic regression, with a decision threshold set \emph{a posteriori} by maximizing the $\Fone$-score on the validation set).
Besides, the structured SVM approach to $\Fone$-measure maximization of \citet{Joachims05}, SVM\textsuperscript{perf}, provides another baseline. 
For completeness, we also report results for non-thresholded cost-sensitive SVMs, non-thresholded cost-sensitive logistic regression, and for the thresholded versions of SVM\textsuperscript{perf}.

Since the practical cost-sensitive algorithms are based on convex surrogate loss optimization \citep{Scott12}, the approximate cost approximation we presented in proposition \ref{prop:approx} will not hold in general.
We call the cost given in proposition \ref{prop:approx} as actual cost and cost used in the practical surrogate loss based algorithm as surrogate cost.
Since there is no one-to-one mapping between actual cost and surrogate cost, in practical implementations we have to iterate over the convex surrogate loss for each value of the actual cost.

SVM and LR differ in the loss they optimize (weighted hinge loss for SVMs, weighted log-loss for LR), and even though both losses are calibrated in the cost-sensitive setting (that is, converging toward a Bayes-optimal classifier as the number of examples and the capacity of the class of function grow to infinity) \citep{Steinwart07}, they behave differently on finite datasets or with restricted classes of functions.
We may also note that asymptotically, the Bayes-classifier for a cost-sensitive binary classification problem is a classifier which thresholds the posterior probability of being class $1$.
Thus, all methods but SVM\textsuperscript{perf} are asymptotically equivalent, and our goal here is to analyze their non-asymptotic behavior on a restricted class of functions.

For each experiment, the training set was split at random, keeping ${1}/{3}$ for the validation set used to select all hyper-parameters, based on the maximization of the $\Fone$-measure on this set.
For datasets that do not come with a separate test set, the data was first split to keep ${1}/{4}$ for test.
All results are averaged over five random splits i.e. hold-out validation with five random splits.
The algorithms have from one to four hyper-parameters: \emph{(i)} all algorithms are run with $L_2$ regularization, with a regularization parameter $C\in\{2^{-6}, 2^{-5}, ..., 2^6\}$; \emph{(ii)} for the cost-sensitive algorithms, the cost for false negatives is chosen in $\{\frac{2-t}{t}, t\in\{0.1,0.2,...,1.9\}\}$ of Proposition~4 \,\footnote{We take $t$ greater than $1$ in case the training asymmetry would be different from the true asymmetry \citep{Bach06}.}; \emph{(iii)} for the thresholded algorithms, the threshold is chosen among all the scores of the validation examples; \emph{(iv)} for kernel based SVM, we used radial basis function (RBF) kernel with $\gamma$ (measure of influence of a single training example) value $\gamma \in\{2^{-6}, 2^{-5}, ..., 2^6\}$.

The library \emph{LIBLINEAR} \citep{fan2008liblinear} was used to implement non-kernel SVMs\footnote{The maximum number of iteration for SVMs was set to $50,\!000$ instead of the default $1,\!000$.} and logistic regression. \emph{LIBSVM} \citep{CC01a} library was used for the kernel SVM.
A constant feature with value $100$ (to simulate an unregularized offset) was added to each dataset.

\subsection{Importance of Thresholding}
\begin{figure}
  \begin{center}
  \begin{tabular}{@{}c@{~}c@{\hspace*{0.1\columnwidth}}c@{~}c@{}}
        & {before thresholding}          & & {after thresholding} \\
    \rotatebox{90}{\hspace*{0.12\columnwidth}$x_{2}$} &
    \includegraphics[width=0.42\columnwidth]{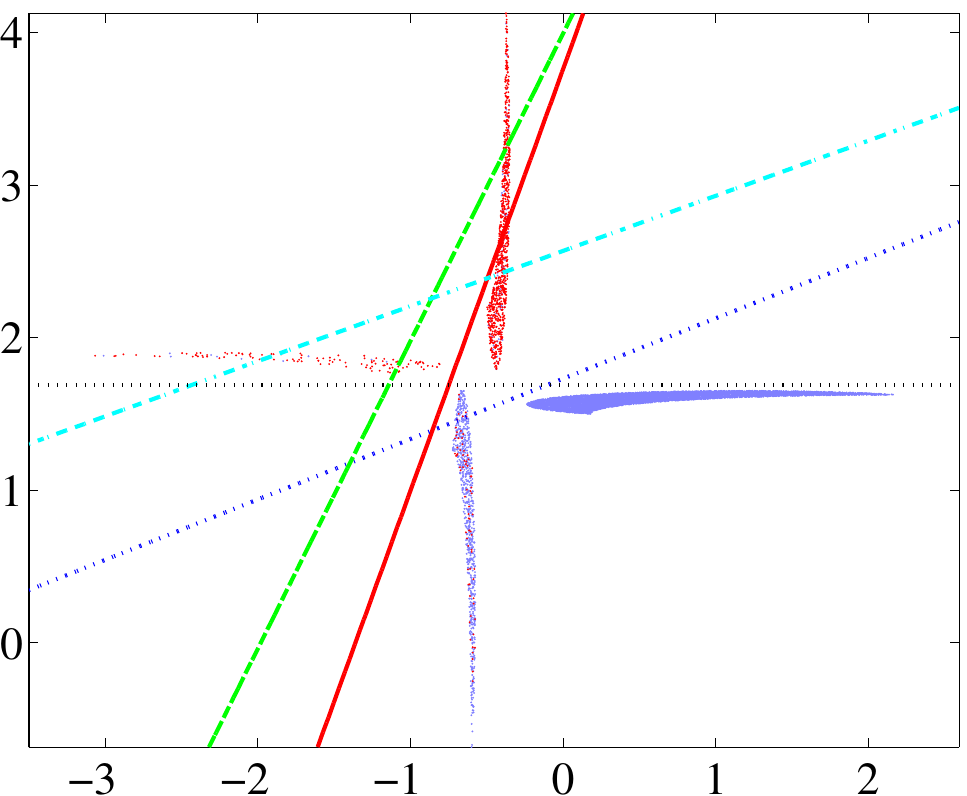} &
    \rotatebox{90}{\hspace*{0.12\columnwidth}$x_{2}$} &
    \includegraphics[width=0.42\columnwidth]{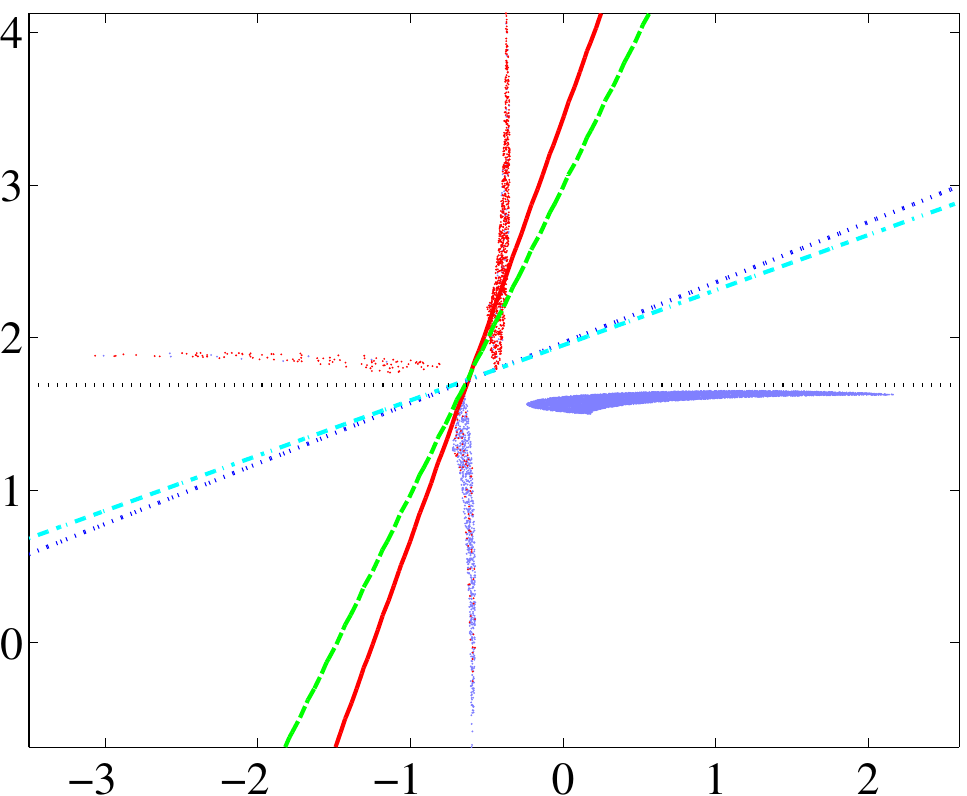} \\[-1ex]
    & $x_{1}$ &  & $x_{1}$ \\[-2ex]
  \end{tabular}
  \end{center}
  \caption{%
  Decision boundaries for the galaxy dataset before and after thresholding
  the classifier scores of
  SVM\textsuperscript{perf} (dotted, blue),
  weighted SVM (dot-dashed, cyan),
  unweighted logistic regression (solid, red),
  and
  weighted logistic regression (dashed, green).
  The horizontal black dotted line is an optimal decision boundary.
  }
  \label{fig:galaxy}
\end{figure}

Although our theoretical developments do not indicate any need to threshold the scores of classifiers, the practical benefits of a post-hoc adjustment of these scores can be important in terms of $\Fone$-measure maximization, as already noted in cost-sensitive learning scenarios \citep{Grandvalet05,Bach06}.
We study the importance thresholding clasification scores \emph{a posteriori} using a didactic data called ``Galaxy''. 
The data can be visualized as given in Figure~\ref{fig:galaxy}.
The data distribution consist in four clusters of 2D-examples, indexed by $z\in\{1,2,3,4\}$, with prior probability $\mu(z=1)=0.01$, $\mu(z=2)=0.1$, $\mu(z=3)=0.001$, and $\mu(z=4)=0.889$, with respective class prior probabilities $\mu(y=1|z=1)=0.9$, $\mu(y=1|z=2)=0.09$, $\mu(y=1|z=3)=0.9$, and $\mu(y=1|z=4)=0$.
``Galaxy'' is an example of highly imbalanced dataset.

We drew a very large sample ($100,\!000$ examples) from the distribution, whose optimal $\Fone$-measure is $67.5\%$.
Without thresholding the scores of the classifiers, the best $\Fone$-measure among the classifiers is $58.0\%$, obtained by cost-sensitive SVM, whereas tuning thresholds enables to reach the optimal $\Fone$-measure for SVM\textsuperscript{perf} and cost-sensitive SVM.
On the other hand, LR is severely affected by the non-linearity of the level sets of the posterior probability distribution, and does not reach this limit (best $\Fone$-measure of $56.5\%$).
Note also that, even with this very large sample size, the SVM and LR classifiers are very different.
This result suggests that thresholding the classification scores \emph{a posteriori} may improve the optimal $F$-scores, especially thresholding the cost-sensitive classifier scores.

\subsection{Binary $F_{\beta}$ and Multilabel $MF_{\beta}$}

\newcommand{\Adult}{{\tt Adult}\xspace}
\newcommand{\Galaxy}{{\tt Galaxy}\xspace}
\newcommand{\Scene}{{\tt Scene}\xspace}
\newcommand{\Siam}{{\tt Siam}\xspace}
\newcommand{\Yeast}{{\tt Yeast}\xspace}
\newcommand{\Rcv}{{\tt RCV1}\xspace}
\newcommand{\Cm}{$C_{\text{\tiny min}}$}
\newcommand{\Fm}{$F_{\text{\tiny max}}$}

The other datasets we use are \Adult, \Rcv, \Scene, \Siam and \Yeast.
In addition, we used a subsample from the \Galaxy data to demonstrate the empirical validity of the algorithm.
\Adult, \Rcv and \Yeast are obtained from the UCI repository\footnote{\url{https://archive.ics.uci.edu/ml/datasets.html}}, and \Scene and \Siam from the \emph{Libsvm} repository\footnote{\url{http://www.csie.ntu.edu.tw/~cjlin/libsvmtools/datasets/multilabel.html}}.
The attributes of the data used in our empirical study are given in Table~\ref{tab:data_desc}.

\begin{table}
\centering
\begin{tabular}{lc*{4}rr@{/}l}
\toprule
Name & Type & Labels & Train & Test & Features & \multicolumn{2}{c}{Label Freq. (\%)} \\
     &      &        &       &      &          & \multicolumn{2}{c}{(min/max)} \\
\midrule
\Adult & binary & 2 & 32,561 & 16,281 & 123 & \multicolumn{2}{c}{--} \\
\Galaxy & binary & 2 & 18,000 & 7,000 & 2 & \multicolumn{2}{c}{--} \\
\Rcv & multilabel & 101 & 23,149 & 10,000 & 47,236 & ~~~0.008 & 46.6\\
\Scene &  multilabel & 6 & 1,211 & 1,196 & 294 & 13.6 & 22.8\\
\Siam &  multilabel & 22 & 21,519 & 7,077 & 30,438 & 1.4 & 59.8\\
\Yeast &  multilabel & 14 & 1,500 & 917 & 103 & 25.2 & 43.0\\
\bottomrule
\end{tabular}
\caption{Dataset Attributes}
\label{tab:data_desc}
\end{table}

The results for binary-$\Fbeta$ and multilabel-macro-F ($\MFbeta$) are reported in Table \ref{tab:tab_bin} and \ref{tab:tab_MF} respectively.
As it is evident from the experimental results, cost-sensitive learning and thresholded cost-sensitive learning give optimal results, whereas other methods performs suboptimally. But the difference between methods is less extreme than on the artificial Galaxy dataset.
The \Adult dataset is an example where all methods perform nearly identical; the surrogate loss used in practice seems unimportant.
On the other datasets, we observe that thresholding has relatively large impact, especially for SVM\textsuperscript{perf} and cost-insensitive classifiers.
The unthresholded and cost-insensitive SVM and LR results are very poor compared to thresholded and cost-sensitive versions.
The cost-sensitive classifiers (thresholded and unthresholded) outperforms all other methods, as suggested by the theory.
Te cost-sensitive SVM is probably the method of choice to optimize binary-$\Fbeta$ or multilabel-macro-F($\MFbeta$) when predictive performance is a must.
On these datasets, thresholded LR still performs reasonably well considering its relatively low computational cost.
In general, on the computational cost front, LR converges faster than SVM or SVM\textsuperscript{perf}.

\begin{table}
  \centering
  \begin{tabular}{l*{12}{c}}
	\toprule
	Baseline
	& \multicolumn{2}{c}{SVM\textsuperscript{perf}} && \multicolumn{4}{c}{SVM} && \multicolumn{4}{c}{LR} \\
	\cmidrule(lr){2-3} \cmidrule(lr){5-8} \cmidrule(lr){10-13}
	Options
	& -- & \scshape t && -- & \scshape t & \scshape cs & \scshape cs{\scriptsize\&}t && -- & \scshape t & \scshape cs & \scshape cs{\scriptsize\&}t \\
	\midrule
    \Adult    & 67.3 & 67.3 && 66.9 & 67.5 & \textbf{67.9} & 67.8 && 65.0 & 67.7 & 67.7 & \bf{67.9} \\
    \Galaxy   & 48.4 & 61.7 && 43.1 & 61.4 & 58.0 & \textbf{62.0} && 35.4 & 51.9 & 41.8 & 56.5 \\
	\bottomrule
  \end{tabular}
  \caption{$F_1$-measures (in \%) for baseline algorithms with their usual settings (--) and different options: T for thresholded classification scores, CS for cost-sensitive training, CS{\small\&}T for cost-sensitive training and thresholded classification scores} \label{tab:tab_bin}
\end{table} 

\begin{table}
  \centering
  \begin{tabular}{l*{12}{c}}
        \toprule
        Baseline
        & \multicolumn{2}{c}{SVM\textsuperscript{perf}} && \multicolumn{4}{c}{SVM} && \multicolumn{4}{c}{LR} \\
        \cmidrule(lr){2-3} \cmidrule(lr){5-8} \cmidrule(lr){10-13}
        Options
        & -- & \scshape t && -- & \scshape t & \scshape cs & \scshape cs{\scriptsize\&}t && -- & \scshape t & \scshape cs & \scshape cs{\scriptsize\&}t \\
        \midrule
        \Rcv    & 44.0 & 52.8 && 46.6 & 54.2 & 50.9 & \textbf{54.5} && 40.9 & 52.9 & 48.5 & 53.3\\
        \Scene  & 68.3 & 69.6 && 66.2 & 69.6 & 69.6 & 69.6 && 67.0 & 69.9 & 69.8 & \textbf{70.1} \\
        \Siam   & 48.2 & 52.8 && 48.1 & 52.4 & 52.7 & \textbf{53.4} && 44.7 & 51.9 & 51.7 & 52.2 \\
        \Yeast  & 46.4 & 46.4 && 39.1 & 46.2 & 47.2 & 46.3 && 38.8 & \textbf{47.4} & \textbf{47.4} & 47.2 \\
        \bottomrule
  \end{tabular}
  \caption{Macro-$F_1$-measures $MF_1$ (in \%) for baseline algorithms with their usual settings (--) and different options: T for thresholded classification scores, CS for cost-sensitive training, CS{\small\&}T for cost-sensitive training and thresholded classification scores}\label{tab:tab_MF}
\end{table}

Table~\ref{tab:tab_MF_kern} presents the optimal $\MFbeta$-measure with kernel SVM.
We used Radial Basis Function (RBF) as the kernel function and trained RBF SVM without a bias term.
Our experiments exemplify our theoretical findings in kernel settings.
In case of $\Scene$, thresholding the cost-sensitive scores marginally improves the $MF_1$-score whereas in case of $\Yeast$ data, cost-sensitive kernel SVM outperforms other methods.
In both cases, thresholding the cost-insensitive scores deteriorates the $MF_1$-scores.

\begin{table}
  \centering
  \begin{tabular}{l*{4}{c}}
    \toprule
    Options  & --      & {\sc t} & {\sc cs} & {\sc cs{\scriptsize \&}t} \\
    \midrule
    \Scene   & 68.9 & 68.3 & 70.5 & \textbf{70.9} \\
    \Yeast   & 48.6 & 48.5 & \textbf{48.8} & 47.9 \\
    \bottomrule
  \end{tabular}
  \caption{Macro-$F_1$-measures $MF_1$ (in \%) for SVM with RBF kernel with their usual settings (--) and different options: T for thresholded classification scores, CS for cost-sensitive training, CS{\small\&}T for cost-sensitive training and thresholded classification scores}
  \label{tab:tab_MF_kern}
\end{table}

\subsection{Multilabel $mF_{\beta}$}

In case of multilabel-micro-F-measure, we compare our algorithm with a commonly used method to find best $\mFbeta$-score suggested by \citet{Fan07}.
In the proposed method, one assumes that an optimal classifier for macro-F-measure is an optimal classifier for micro-F-measure.
Hence, the micro-F-score corresponds to optimal macro-F-score is deemed as the optimal micro-F-score.
We compare our algorithm for micro-F-score against the micro-F-score corresponds to the optimal macro-F-score obtained by running binary relevance as explained in section~\ref{sec:mullab_fbeta}. 

\begin{table}
  \centering
  \begin{tabular}{l@{\ }c*{13}{c}}
        \toprule
        \multicolumn{2}{l}{Baseline}
        & \multicolumn{2}{c}{SVM\textsuperscript{perf}} && \multicolumn{4}{c}{SVM} && \multicolumn{4}{c}{LR} \\
        \cmidrule(lr){3-4} \cmidrule(lr){6-9} \cmidrule(lr){11-14}
        \multicolumn{2}{l}{Options}
        & -- & \scshape t && -- & \scshape t & \scshape cs & \scshape cs{\scriptsize\&}t && -- & \scshape t & \scshape cs & \scshape cs{\scriptsize\&}t \\
        \midrule
        \multicolumn{1}{l}{\multirow{2}{*}{\Rcv}} & \Cm & 48.2 & 49.6 && 47.6 & 49.7 & 49.9 & \textbf{50.2} && 46.3 & 49.8 & 49.9 & 49.9 \\
        \multicolumn{1}{}{} & \Fm & 42.8 & 44.7 && 47.6 & 44.1 & 49.2 & 44.2 && 46.4 & 44.3 & 49.3 & 44.5 \\
        \hline
        \multicolumn{1}{l}{\multirow{2}{*}{\Scene}} & \Cm & 66.7 & 68.5 && 65.4 & 68.7 & 68.8 & 68.6 && 66.6 & 69.2 & 68.6 & \textbf{69.4} \\
        \multicolumn{1}{}{} & \Fm & 66.6 & 68.3 && 65.2 & 68.3 & 68.3 & 68.3 && 66.4 & 69.2 & 68.6 & 68.8 \\
        \hline
        \multicolumn{1}{l}{\multirow{2}{*}{\Siam}} & \Cm & 59.2 & 62.5 && 60.3 & 62.2 & \textbf{62.6} & 62.5 && 60.2 & 62.4 & 62.0 & 62.3 \\
        \multicolumn{1}{}{} & \Fm & 59.2 & 62.0 && 60.1 & 62.0 & 62.3 & 62.2 && 59.0 & 61.8 & 61.9 & 62.0 \\
        \hline
        \multicolumn{1}{l}{\multirow{2}{*}{\Yeast}} & \Cm & 61.8 & 65.1 && 64.1 & 64.8 & \textbf{65.6} & 65.2 && 63.3 & 64.9 & 65.3 & 64.9 \\
        \multicolumn{1}{}{} & \Fm & 60.2 & 60.2 && 60.6 & 59.3 & 60.7 & 61.2 && 63.2 & 59.8 & 61.0 & 60.9 \\
        \bottomrule
  \end{tabular}
  \caption{Micro-$F_1$-measures $mF_1$ (in \%) for for baseline algorithms with their usual settings (--) and different options: T for thresholded classification scores, CS for cost-sensitive training, CS{\small\&}T for cost-sensitive training and thresholded classification scores.
Two optimization strategies are compared:
{\Cm} for $mF_1$ by proposed algorithm and {\Fm} for $mF_1$ corresponding to optimal $MF_1$}
\label{tab:tab_mF}
\end{table}

Table~\ref{tab:tab_mF} contains the multilabel-micro-F ($\mcFbeta$) results for the multilabel datasets.
The results clearly demonstrates that selecting micro-F corresponds to maximal macro-F (correspond to \Fm in table) always return suboptimal results.
So in practice, algorithms based on per-label macro-F optimization should be avoided for micro-F optimization.
In case of micro-F, effect due to thresholding is not very significant, except for $\Rcv$ data.
The unthresholded classifiers performs nearly as good as the thresholded versions.
This is true for SVM\textsuperscript{perf} also.
As suggested by theory, cost-sensitive classification is the preferred method to optimize multilabel-micro-F.
Here also, thresholded LR can be considered as an alternate option considering the computational cost.

\begin{table}
  \centering
  \begin{tabular}{l@{\ }c*{4}{c}}
    \toprule
    \multicolumn{2}{l}{Options}
    & --      & {\sc t} & {\sc cs} & {\sc cs{\scriptsize \&}t} \\
    \midrule
    \multicolumn{1}{l}{\multirow{2}{*}{\Scene}} & \Cm & 67.2 & 67.1 & \textbf{67.5} & 67.1 \\
    \multicolumn{1}{}{} & \Fm & 67.0 & 67.0 & 67.2 & 67.4 \\
    \midrule
    \multicolumn{1}{l}{\multirow{2}{*}{\Yeast}} & \Cm  & 65.9 & 66.3 & 66.3 & \textbf{66.6} \\
    \multicolumn{1}{}{} & \Fm & 59.4 & 62.9 & 59.9 & 63.5 \\
    \bottomrule
  \end{tabular}
  \caption{$mF_1$  for SVM with RBF kernel with their usual settings (--) and different options: T for thresholded classification scores, CS for cost-sensitive training, CS{\small\&}T for cost-sensitive training and thresholded classification scores.
{\Cm} for $mF_1$ by proposed algorithm and {\Fm} for $mF_1$ corresponding to optimal $MF_1$}
  \label{tab:tab_mF_kern}
\end{table}

Table~\ref{tab:tab_mF_kern} presents the optimal $\mcFbeta$-measure with RBF kernel SVM.
Similar to the $\MFbeta$ results, thresholding the cost-sensitive score gives better $mFbeta$ results for kernel SVM.

\subsection{Cost Space Search Overhead}
Since the actual cost associated misclassification differs from the cost associated with surrogate loss, it introduces an extra loop in our algorithm. Hence searching for optimal cost vector in the  discretized cost interval might not be a good idea, especially when the value of $\Lmu$ is large.
Here we do an empirical analysis of the functional dependencies between the actual cost and corresponding $F$-measure, and devise an improved version of the algorithms discussed in Section~\ref{sec:beyond_fbeta}.

\begin{figure}
\centering
\includegraphics[scale=0.51,trim = 0cm 15cm 0cm 15cm]{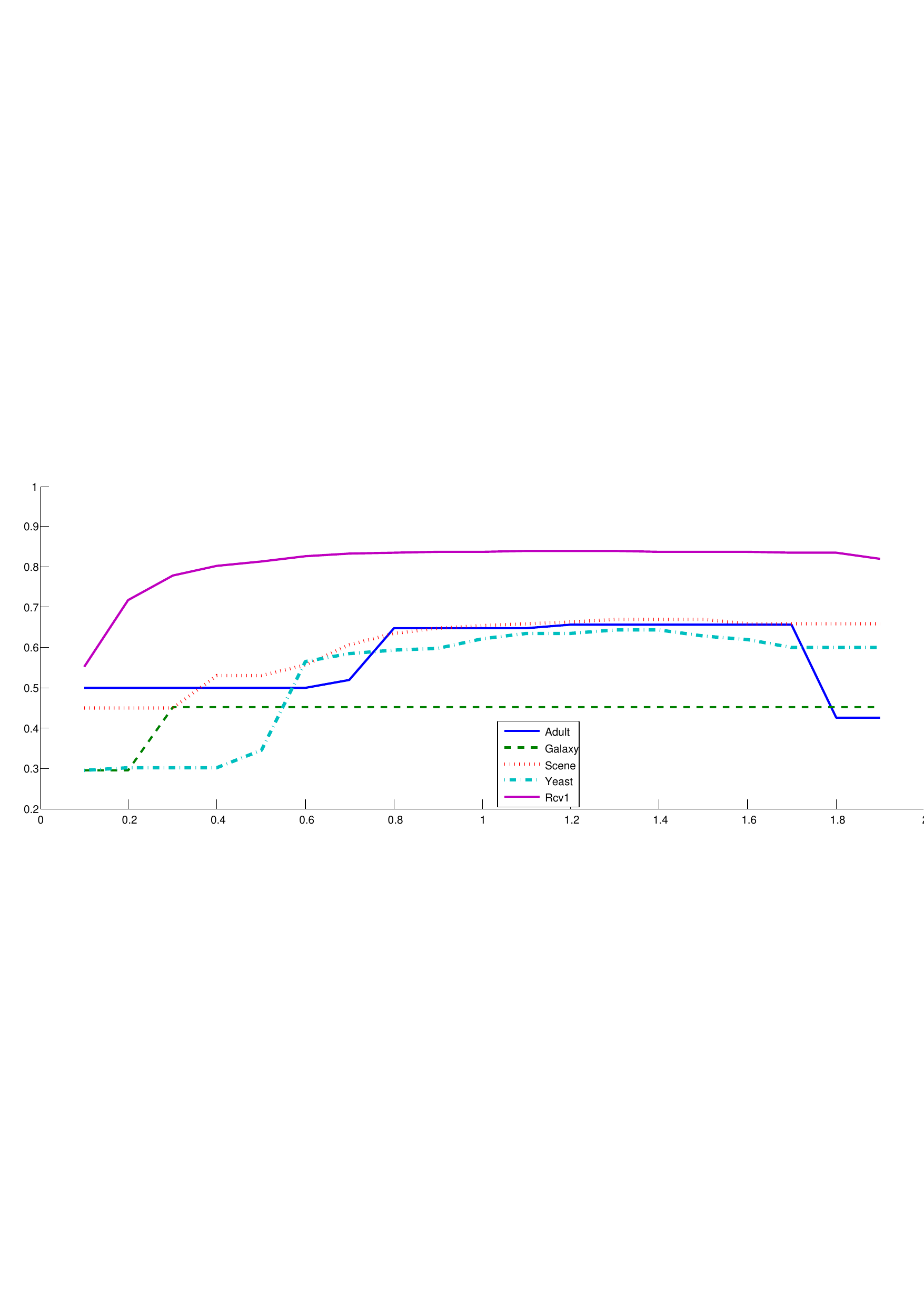}
\caption{Plot of micro-$F$-measure against false negative cost}
\label{fig:cost_mF}
\end{figure}

Figure~\ref{fig:cost_mF} contains the plot of micro-$F$-measure against false negative cost.
From the plot, it is evident that micro-$F$-measure is a quasi-concave function of false negative cost.
A function is quasi-concave, if every superlevel set of the function is convex \citep{Boyd:2004:CO:993483}.
Formally, a function $g:\domF\subset\Re^d\rightarrow \Re$, is quasi-concave if $ \{x \in \domF ~|~ g(x) \ge a\}$ is convex.
It can be verified from the plot that superlevel sets are convex.
Bracketing methods \citep{NRE07} are extensively used to find global maxima of unimodal functions like quasi-concave function.
We will not be able to use the exact bracketing algorithm to find the optimal cost, since it requires the knowledge of \emph{error profile} associated with each value of $F$-measure).
But we can use the idea of bracketing to limit the discretization interval.

Here, we find three points ($p,q,r$), such that $g(p)<g(q)>g(r)$, then instead of discretizing the whole interval, we can limit the discretization only to the sub-interval $(p,r)$.
We start with two intervals defined by the three points: start of the interval ($0$), median of the interval ($\frac{1+\beta^2}{2}$) and the end of the interval ($1+\beta^2$).
Then we search for the triplets ($p,q,r$) of given minimum sub-interval size inside the two intervals.
In the simplest case, we find  $F$-measure values corresponding to five points, two start points, midpoint ($\frac{1+\beta^2}{2}$) and two midpoints of the intervals ($0,\frac{1+\beta^2}{2}$) and ($\frac{1+\beta^2}{2},1+\beta^2$).
Since the function is quasi-concave, the global maxima can be either on the mid point or on left or right  of the mid point.
Depending up on the $F$-measure values at the five points, we can limit the discretization only to one half.
This way we can reduce the discretization space at least by half.

\section{Conclusion}
\label{sec:discus}
We presented an analysis of $F$-measures, leveraging the property of pseudo-linearity of specific notions of $F$-measures to obtain a strong non-asymptotic reduction to cost-sensitive classification.
The results hold on any dataset, for any class of function and on any data distribution assumptions (label dependent or label independent).
We suggested algorithms for $F$-measure optimization based on minimizing the total misclassification cost of the cost-sensitive classification.
We demonstrated experiments on linear classifiers, showing the theoretical interest of using cost-sensitive classification algorithms rather than probability thresholding.
It is also shown that for $F$-measure maximization, thresholding even the cost-sensitive
algorithms helps to achieve good performances.

Empirically and algorithmically, we only explored the simplest case of our result ($\Fbeta$-measure in binary classification and macro-$\Fbeta$-measure and micro-$\Fbeta$-measure in multilabel classification), but much more remains to be done. Algorithms for the optimization of the non-pseudo-linear notions of $F$-measures like instance-wise-$\Fbeta$-measure in multilabel classification received interest recently as well \citep{Dembczynski11,Cheng12}, but are for now limited. We also believe that our result can lead to progresses towards optimizing the micro-$\Fbeta$ measure in multiclass classification.

\acks{This work was carried out and funded in the framework of the Labex MS2T. It was
supported by the Picardy Region and the French Government, through the program
``Investments for the future'' managed by the National Agency for Research
(Reference ANR-11-IDEX-0004-02)}



\appendix
\section{Proofs of Propositions and Corollaries}
\label{app:thm_psd_lin}

\noindent {\bf Proposition \ref{prop:fraclin}~}{\em
A linear-fractional function
$F:\domF\subseteq\Re^d\rightarrow \Re$
is the ratio of linear functions
$F(\br) = \frac{\alpha_0 + \dotp{\boldsymbol{\gamma}}{\br}}{\alpha_1 + 
\dotp{\boldsymbol{\delta}}{\br}}$.
A non-constant linear-fractional function is pseudo-linear on the open half-space
$\domF = \left\{\br\in\Re^d|\alpha_1+\dotp{\boldsymbol{\delta}}{\br}  > 0\right\}$.
}

\begin{rproof}
A linear-fractional function $F:\br\in \Re^d\mapsto \frac{\alpha_0 + \dotp{\boldsymbol{\gamma}}{\br}}{\alpha_1 + \dotp{\boldsymbol{\delta}}{\br}}\enspace , ~\alpha_1+\dotp{\boldsymbol{\delta}}{\br}  > 0$ is pseudo-linear.

\begin{equation*}
\begin{split}
F(\br)\leq t \Leftrightarrow  & \alpha_0 + \dotp{\boldsymbol{\gamma}}{\br} \leq t(\alpha_1 + \dotp{\boldsymbol{\delta}}{\br}) \\
\Rightarrow & (\alpha_0 - t\alpha_1) + \dotp{\boldsymbol{\gamma}-t\boldsymbol{\delta}}{\br} \leq 0 \\
\end{split}
\end{equation*}
Now reversing the inequality, we obtain;

\begin{equation*}
F(\br)\geq t \Leftrightarrow (\alpha_0 - t\alpha_1) + \dotp{\boldsymbol{\gamma}-t\boldsymbol{\delta}}{\br} \geq 0 \\
\end{equation*}
Above equations represent open hyperplanes.

\begin{equation*}
\nabla F(\br) = \frac{(\alpha_1 + \dotp{\boldsymbol{\delta}}{\br})\boldsymbol{\gamma} - (\alpha_0 + \dotp{\boldsymbol{\gamma}}{\br})\boldsymbol{\delta}}{(\alpha_1 + \dotp{\boldsymbol{\delta}}{\br})^2} \neq 0
\end{equation*}

The gradient term is constant if $\delta$ and $\gamma$ are propotional and non-zero otherwise. The above conditions confirm the requirements for the pseudo-linearity given in Theorem~\ref{thm:qualif} and hence the result. 
\hfill $\Box$
\end{rproof}

\noindent {\bf Proposition \ref{prop:weightedsum}~}{\em
Let ${\displaystyle \maxFmuH = \max_{\br\in\rpfhypo} \Fmu{\br}}$, we have:~
$\displaystyle \br^\star \in \argmin_{\br\in\rpfhypo} \dotp{\bamu\big(\maxFmuH\big)}{\br}~\Leftrightarrow~\Fmu{\br^\star}=\maxFmuH$ .}
\begin{rproof}
Let $\br^\star\in\argmax_{\brprime\in\rpfhypo} \Fmu{\brprime}$, and let
$\ba^\star = \bamu{\Fmu(\br^\star)}=\bamu\big(\maxFmuH\big) $.
We first notice that pseudo-linearity implies that the set of $\br\in\domFmu$ such that $\dotp{\ba^\star}{\br} = \dotp{\ba^\star}{\br^\star}$ corresponds to the level set $\{\br\in\domFmu|\Fmu{\br}= \Fmu{\br^\star}=\maxFmuH\}$. Thus, we only need to show that $\br^\star$ is a minimizer of $\brprime\mapsto\dotp{\ba^\star}{\brprime}$ in $\rpfhypo$. To see this, we notice that pseudo-linearity of $\Fmu$ (see Theorem~\ref{thm:qualif}) implies
\begin{equation*}
  \forall \brprime\in\domFmu, ~ \Fmu{\br^\star} \ge \Fmu{\brprime} \Rightarrow \dotp{\ba^\star}{\br^\star} \le \dotp{\ba^\star}{\brprime}
  \enspace,
\end{equation*}
and since $\br^\star$ maximizes $\Fmu$ in $\rpfhypo$, we get $\br^\star \in \argmin_{\brprime\in\rpfhypo} \dotp{\ba^\star}{\brprime}$ .
\hfill $\Box$
\end{rproof}

\noindent {\bf Proposition \ref{prop:approx}~}{\em
Let $\varepsilon_0\geq 0$ and $\varepsilon_1\geq 0$, and assume that there exists
$\Lmu>0$ such that for all $\br, \brprime\in\rpfhypo$ satisfying $\Fmu{\brprime}>\Fmu{\br}$, we have:
\begin{equation}
  \label{eq:bamucool}
  \Fmu{\brprime}-\Fmu{\br}\leq \Lmu\dotp{\bamu{\Fmu(\brprime)}}{\br-\brprime}
  \enspace.  
\end{equation}
Then, let us take $\br^\star\in\argmax_{\brprime\in\rpfhypo} \Fmu{\brprime}$, and denote $\ba^\star=\bamu{\Fmu(\br^\star)}$. Let furthermore $\bgg\in\Re_+^d$ and $h\in\hyposp$ satisfying the following conditions:
\begin{center}
{(i)} $\norm{\bgg-\ba^\star}\leq \varepsilon_0$
\enspace,
\hspace{2.5cm} {(ii)} $\displaystyle \dotp{\bgg}{\br} \leq \min_{\brprime\in\rpfhypo} \dotp{\bgg}{\brprime}+\varepsilon_1$
\enspace.
\end{center}
We have: $\forall \br \in \rpfhypo, ~ \Fmu{\br}\geq \Fmu{\br^\star} - \Lmu \cdot (2\varepsilon_0M+\varepsilon_1)$
\enspace,
\enspace where $\displaystyle M=\max_{\brprime\in\rpfhypo}{\norm{\brprime}}$
\enspace. 
}

\begin{rproof}
Let $\brprime \in \rpfhypo$, we can write
$
\dotp{\bgg}{\brprime} = \dotp{\ba^\star}{\brprime} + \dotp{\bgg - \ba^\star}{\brprime}
$.
Applying Cauchy-Schwarz inequality and condition {\em (i)}, we get
\begin{align*}
  \dotp{\bgg}{\brprime} 
  & \leq \dotp{\ba^\star}{\brprime} + \norm{\bgg-\ba^\star}\norm{\brprime} \\
  & \leq \dotp{\ba^\star}{\brprime} + \varepsilon_0 M 
  \enspace.
\end{align*}
In particular, we have:
\begin{align}
  \min_{\brprime\in\rpfhypo} \dotp{\bgg}{\brprime} 
  & \leq  \min_{\brprime\in\rpfhypo} \dotp{\ba^\star}{\brprime}+\varepsilon_0M \nonumber \\
  & \leq  \dotp{\ba^\star}{\br^\star} + \varepsilon_0 M \label{eq:remark1}
  \enspace,
\end{align}
since $\br^\star \in \argmin_{\brprime\in\rpfhypo} \dotp{\ba^\star}{\brprime}$ as shown in Proposition~\ref{prop:weightedsum}.

Similarly, we have $\dotp{\ba^\star}{\br} = \dotp{\bgg}{\br} + \dotp{\ba^\star - \bgg}{\br}$;
applying Cauchy-Schwarz and conditions {\em (i)} and {\em (ii)}, we have:
\begin{align}
  \forall \br \in \rpfhypo, ~ \dotp{\ba^\star}{\br} 
  & \leq \dotp{\bgg}{\br} + \norm{\ba^\star-\bgg}\norm{\br} \nonumber \\
  & \leq \dotp{\bgg}{\br} + \varepsilon_0 M \nonumber \\
  & \leq \min_{\brprime\in\rpfhypo} \dotp{\bgg}{\brprime}+ \varepsilon_1 + \varepsilon_0 M \label{eq:remark2}
  \enspace.
\end{align}
Combining Inequalities~\eqref{eq:remark1} and \eqref{eq:remark2}, we get
\begin{align*}
  \forall \br \in \rpfhypo, ~ 
  & \dotp{\ba^\star}{\br} 
  \leq \dotp{\ba^\star}{\br^\star} + \varepsilon_1 + 2 \varepsilon_0 M \\
  \forall \br \in \rpfhypo, ~ 
  & \dotp{\ba^\star}{\br-\br^\star} 
  \leq \varepsilon_1 + 2 \varepsilon_0 M   \enspace,
\end{align*}
and the final result follows from Assumption~\eqref{eq:bamucool}. \hfill $\Box$
\end{rproof}


\noindent {\bf Proposition \ref{prop:fscoresforreal}~}{ \em
$\Fbeta$-measures defined in Section \ref{sec:bin_fbeta} satisfy the conditions of Proposition \ref{prop:approx} with:
\begin{equation*}
{\scriptstyle (binary)~~\Fbeta:} \hspace{2.1cm} \!\Lmu=\frac{1}{\beta^2\Pmu_1} \hspace{1.1cm} \text{~~~and~~} \bamu:t\in[0,1]\mapsto(1+\beta^2-t,t,0,0)\enspace .\hspace{1.6cm}
\end{equation*} }

\begin{rproof}
Since $\Fbeta$ is linear-fractional as a function of the error profile, it is pseudo-linear on the open convex set $\{\br\in\Re^d| (1+\beta^2)P_1-e_1+e_2>0\}$ (i.e. when the denominator is strictly positive). Moreover, for every set of classifiers $\hyposp$, we have $\rpfhypo\subseteq\domFbeta_0=[O,\Pmu_1]\times[0,1-\Pmu_1]\times[1-\Pmu_1]\times[1,\Pmu_1]$.

Now, by the definition of $\Fbeta$, we have
\begin{equation*}
\forall \br\in\domFbeta_0, \Fbeta{\br}\leq t~~~ \Leftrightarrow ~~~(1+\beta^2-t)\br_1+t\br_2+(1+\beta^2)\Pmu_1(t-1)\geq 0\enspace ,
\end{equation*}
and the equation still holds by reversing the inequalities. We thus have that
$\bamu(t) = (1+\beta^2-t,t,0,0)$ satisfy the condition of Theorem \ref{thm:qualif} (with $\bbb(t) = (1+\beta^2)\Pmu_1(t-1)$).\\

We now show that the condition of Equation \ref{eq:bamucool} is satisfied for $\bamu(t) = (1+\beta^2-t,t,0,0)$ and all $\br, \br' \in\domFbeta_0$ by taking $\Lmu =\frac{1}{\beta^2\Pmu_1}$. To that end, let $\br$ and $\brprime$ in $\rpfhypo$ and $t$ and $t'$ in $\Re$ such that $t'=\Fbeta(\brprime)>\Fbeta(\br)=t$. Denote by $\varepsilon$ the quantity $\dotp{\bamu(t')}{\br-\brprime}$. Note that $\varepsilon>0$ and that:
\begin{equation*}
\begin{array}{r c c c c c c c c l}
0& =& \dotp{\bamu(t)}{\br}& +&\bbb(t)& = (1+\beta^2-t)\br_1 &+&t\br_2&+&(1+\beta^2)\Pmu_1(t-1)\\[.2cm]
0& =& \dotp{\bamu(t')}{\brprime}& +& \bbb(t')& = (1+\beta^2-t')\brprime_1 &+&t'\brprime_2&+&(1+\beta^2)\Pmu_1(t'-1)\\[.2cm]
\varepsilon& =& \dotp{\bamu(t')}{\br-\brprime}& & &= (1+\beta^2-t')\br_1 &+&t'\br_2&+&(1+\beta^2)\Pmu_1(t'-1)
\end{array}
\end{equation*}
where the first two equalities are given by the definition of hyperplane corresponds to $\Fbeta(\br)=t$ and $\Fbeta(\brprime)=t'$, and the last one is obtained from the definition of  $\dotp{\bamu(t')}{\br-\brprime}$.
Taking the difference of the third and first equality, we obtain:
\begin{equation*}
\varepsilon = (t-t')\br_1+(t'-t)\br_2+(1+\beta^2)\Pmu_1(t'-t)
\end{equation*}
From which we get, since $(1+\beta^2)\Pmu_1-\br_1+\br_2>0$ for $\br\in\domFbeta_0$:
\begin{equation*}
\Fbeta(\brprime)-\Fbeta(\br)=t'-t=\varepsilon\big((1+\beta^2)\Pmu_1-\br_1+\br_2\big)^{-1}\leq \frac{\varepsilon}{\beta^2\Pmu_1}\enspace ,
\end{equation*}
because $\beta^2\Pmu_1$ the minimum of $(1+\beta^2)\Pmu_1-\br_1+\br_2$ on $\domFbeta_0$ (taking $\br_1=\Pmu_1$ and $\br_2=0$). We obtain the result since $\varepsilon = \dotp{\bamu(t')}{\br-\brprime}$ by definition. \hfill $\Box$
\end{rproof}


\noindent {\bf Corollary \ref{coro:foptcost}~} {\em
For the $\Fone$-measure, the optimal classifier is the solution to the cost-sensitive binary classifier with costs $\big(1-\frac{\maxFmuH}{2},\frac{\maxFmuH}{2}\big)$
}

\begin{rproof}
From Proposition 4, by putting $\beta = 1$, we have
\begin{equation*}
(2-\maxFmuH)\br_1 + \br_2\maxFmuH+ 2\Pmu_1(\maxFmuH-1) \geq 0
\end{equation*}
dividing by 2, we get
\begin{equation*}
(1-\frac{\maxFmuH}{2})\br_1 + \br_2\frac{\maxFmuH}{2}+ \Pmu_1(\maxFmuH-1) \geq 0
\end{equation*}
Cost vector, $\bamu(t)$, according to Theorem~\ref{thm:qualif} is ($1-\frac{\maxFmuH}{2}, \frac{\maxFmuH}{2}$). \hfill $\Box$
\end{rproof}


\noindent {\bf Proposition \ref{prop:mfscoresforreal}~}{\em
multilabel micro-$F$($\mFbeta$) measures defined in Section~\ref{sec:mullab_fbeta} satisfy the conditions of Proposition \ref{prop:approx} with:
\begin{equation*}
{\scriptstyle (multilabel \text{--} micro)~~\mFbeta:} \hspace{0.41cm} \,\Lmu=\frac{1}{\beta^2\sum_{k=1}^\nL \Pmuk} \hspace{0.15cm} \text{~~and~~} \bamu_i(t)=\begin{cases}1+\beta^2-t& \text{~if $i$ is odd}\\ t &\text{~if $i$ is even}\end{cases}\enspace .\hspace{1.75cm}
\end{equation*}
}

\begin{rproof}
\begin{equation*}
\begin{split}
\mFbeta{\br} \le t
\implies \frac{(1+\beta^2)\sum_{k=1}^\nL(\Pmuk-\br_{2k-1})}{(1+\beta^2)\sum_{k=1}^\nL\Pmuk +\sum_{k=1}^\nL(\br_{2k}- \br_{2k-1})} \le t \\
\implies (1+\beta^2-t)\sum_{k=1}^\nL \br_{2k-1} + t \sum_{k=1}^\nL \br_{2k} + (1+\beta^2)(t-1)\sum_{k=1}^\nL \Pmuk \ge 0
\end{split}
\end{equation*}

Thus, we have that
\begin{equation*}
\bamu_i(t)=\begin{cases}1+\beta^2-t& \text{~if $i$ is odd}\\ t &\text{~if $i$ is even}\end{cases}
\end{equation*}

Following the same arguments as in Proposition: 4, we get
\begin{equation*}
\mFbeta(\brprime)-\mFbeta(\br)=t'-t=\varepsilon\bigg[(1+\beta^2)\sum_{k=1}^\nL \Pmuk-\sum_{k=1}^\nL \br_{2k-1}+\sum_{k=1}^\nL \br_{2k}\bigg]^{-1}\leq \frac{\varepsilon}{\beta^2\sum_{k=1}^\nL\Pmuk}\enspace ,
\end{equation*}
because $\beta^2\sum_{k=1}^\nL\Pmuk$ the minimum of $(1+\beta^2)\sum_{k=1}^\nL \Pmuk-\sum_{k=1}^\nL \br_{2k-1}+\sum_{k=1}^\nL \br_{2k}$ in the respective domain (taking $\br_{2k-1}=\Pmuk$ and $\br_{2k}=0$). We obtain the result since $\varepsilon = \dotp{\bamu(t')}{\br-\brprime}$ by definition. \hfill $\Box$
\end{rproof}


\noindent{ \bf Proposition \ref{prop:mcfscoresforreal}~} {\em
multiclass micro-$F$($\mcFbeta$) defined in Section~\ref{sec:mulclass_fbeta} satisfy the conditions of Proposition \ref{prop:approx} with:
\begin{equation*}
{\scriptstyle (multiclass \text{--} micro)~~\mcFbeta:} \hspace{0.37cm} \!\Lmu=\frac{1}{\beta^2(\Pmupos)} \hspace{0.31cm} \text{~~and~~} \bamu_i (t)=\begin{cases}1+\beta^2-t& \text{~if $i$ is odd and $i\neq 1$}\\ t &\text{~if $i=1$}\\0&\text{~otherwise}\end{cases}\enspace .\hspace{0.2cm}
\end{equation*}
}

\begin{rproof}
\begin{equation*}
{\scriptstyle (multiclass \text{--} micro)~~\mcFbeta:} \hspace{0.37cm} \!\Lmu=\frac{1}{\beta^2(\Pmupos)} \hspace{0.31cm} \text{~~and~~} \bamu_i (t)=\begin{cases}1+\beta^2-t& \text{~if $i$ is odd and $i\neq 1$}\\ t &\text{~if $i=1$}\\0&\text{~otherwise}\end{cases}\enspace .\hspace{0.2cm}
\end{equation*}

\begin{equation*}
\begin{split}
\mcFbeta{\br} \le t
\implies \frac{(1+\beta^2)(\Pmupos - \sum_{k=2}^\nL \br_{2k-1})}{(1+\beta^2)(\Pmupos) - \sum_{k=2}^\nL\br_{2k-1}+\br_1} \le t \\
\implies (1+\beta^2-t)\sum_{k=2}^\nL \br_{2k-1} + t \br_{1} + (1+\beta^2)(t-1)(1- \Pmu_1) \ge 0
\end{split}
\end{equation*}

Thus, we have that
\begin{equation*}
\bamu_i (t)=\begin{cases}1+\beta^2-t& \text{~if $i$ is odd and $i\neq 1$}\\ t &\text{~if $i=1$}\\0&\text{~otherwise}\end{cases}
\end{equation*}

Following the same arguments as in Proposition:4, we get
\begin{equation*}
\mcFbeta(\brprime)-\mcFbeta(\br)=t'-t=\varepsilon\bigg[(1+\beta^2)(1-\Pmu_1)-\sum_{k=2}^\nL \br_{2k-1}+ \br_{1}\bigg]^{-1}\leq \frac{\varepsilon}{\beta^2(1-\Pmu_1)}\enspace ,
\end{equation*}
because $\beta^2(1-\Pmu_1)$ the minimum of $(1+\beta^2)(1-\Pmu_1)-\sum_{k=2}^\nL \br_{2k-1}+ \br_{1}$ in the respective domain (taking $\sum_{k=2}^\nL \br_{2k-1} = 1-\Pmu_1$ and $\br_{1}=0$). We obtain the result since $\varepsilon = \dotp{\bamu(t')}{\br-\brprime}$ by definition. \hfill $\Box$
\end{rproof}

\vskip 0.2in
\bibliography{f1}

\end{document}